\documentclass{article}

\usepackage{arxiv}
\usepackage{natbib}

\usepackage[utf8]{inputenc} 
\usepackage[T1]{fontenc}    
\usepackage{hyperref}       
\usepackage{url}            
\usepackage{booktabs}       
\usepackage{amsfonts}       
\usepackage{nicefrac}       
\usepackage{microtype}      
\usepackage{algorithm}      
\usepackage[noend]{algpseudocode}
\usepackage{amsmath}
\usepackage{mathtools}
\usepackage{graphicx}
\usepackage{chngcntr}       
\graphicspath{ {./images/} }

\usepackage[dvipsnames]{xcolor}
\usepackage{soul}
\usepackage{upgreek}

\newcommand{\new}[1]{{ \color{black}{#1}}}
\newcommand{\review}[1]{{\color{black}{#1}}}

\renewcommand*{\thefootnote}{\fnsymbol{footnote}}
\newcommand\blfootnote[1]{%
  \begingroup
  \renewcommand\thefootnote{}\footnote{#1}%
  \addtocounter{footnote}{-1}%
  \endgroup
}

\def\NoAlgNumber#1{{\def\alglinenumber##1{}\State #1}\addtocounter{ALG@line}{-1}}
\newcommand{\matname}{B}

\title{The Dual PC Algorithm and the Role of Gaussianity for Structure Learning of Bayesian Networks}
\shorttitle{The Dual PC Algorithm for Structure Learning} 

\author{
 Enrico Giudice \\
  Dep.\ of Mathematics and Computer Science\\
  University of Basel,
  Basel, Switzerland \\
  \texttt{enrico.giudice@unibas.ch} \\
   \And
 Jack Kuipers \\
  Dep.\ of Biosystems Science and Engineering\\
  ETH Zurich,
  Basel, Switzerland \\
  \texttt{jack.kuipers@bsse.ethz.ch} \\
  \And
 Giusi Moffa \\
  Dep.\ of Mathematics and Computer Science,
  University of Basel, 
  Basel, Switzerland \\
  and Division of Psychiatry,
  University College London, 
  London, UK \\
  \texttt{giusi.moffa@unibas.ch} \\
}

\begin{document}
\maketitle
\begin{abstract}
Learning the graphical structure of Bayesian networks is key to describing data-generating mechanisms in many complex applications but poses considerable computational challenges. Observational data can only identify the equivalence class of the directed acyclic graph underlying a Bayesian network model, and a variety of methods exist to tackle the problem. Under certain assumptions, the popular PC algorithm can consistently recover the correct equivalence class by reverse-engineering the conditional independence (CI) relationships holding in the variable distribution. The dual PC algorithm is a novel scheme to carry out the CI tests within the PC algorithm by leveraging the inverse relationship between covariance and precision matrices. By exploiting block matrix inversions we can \review{also} perform tests on partial correlations of complementary (or dual) conditioning sets. The multiple CI tests of the dual PC algorithm proceed by first considering marginal and full-order CI relationships and progressively moving to central-order ones. Simulation studies show that the dual PC algorithm outperforms the classic PC algorithm both in terms of run time and in recovering the underlying network structure,\new{even in the presence of deviations from Gaussianity. Additionally, we show that the dual PC algorithm applies for Gaussian copula models, and demonstrate its performance in that setting.}
\end{abstract}


\vspace{0.5cm}
\section{Introduction}
\label{intro}
Characterizing the relationships among a set of random variables constitutes a central question in statistics. Probabilistic graphical models provide a compact way of describing a joint probability distribution and enable inference about features of interest. \review{Compared to typical co-occurrence networks probabilistic graphical models have the advantage of accounting for higher order dependencies between the variables, going beyond the description of simple pairwise associations \citep{angelopoulosEtAl2022}.} Bayesian networks (BNs) are one such class of models whose representation employs directed acyclic graphs (DAGs) to describe the set of probabilistic dependencies among variables \citep{vstruc}. Every node on the graph represents a variable with edges between pairs of nodes encoding conditional independence relations among the variables. \review{Technically, the joint probability distribution decomposes into a product of terms, each typically involving only a few variables. 

By aiming to imitate real-world data generating processes \citep{pearl2000}, Bayesian networks have found applications in a wide array of diverse fields, such as characterising gene expression patterns in genomics \citep{friedmangene2000, friedmancell2004, genereview}, capturing the interplay between mutated genes and possibly uncovering novel genomic-based patient stratifications to inform the development of personalised treatments \citep{naturegene}, performing text classification \citep{textclas, textclas2}, in the social sciences \citep{elwertsocial, creditr} or describing Covid infection rates in epidemiology\citep{covid}. 

Under certain assumptions Bayesian networks may also depict the probabilistic relationships ensuing from causal links between a set of variables \citep{vstruc}. Attempting to learn the underlying graphical structure describing causal mechanisms compatible with the observations constitutes a fundamental step in causal discovery, with prominent examples of applications in genomic \citep{ida} and psychology \citep{bully, moffa_kuipers_2021, bnsparanoia}.} This variety of applications has driven interest in Bayesian network inference, especially in high-dimensional settings where prior knowledge of the DAG structure is unrealistic. When BNs represent causally induced conditional independencies, the edges in the DAG indicate direct causal effects from one variable to another.

\subsection{Structure Learning}
\label{struc}
\review{Let $\textbf{X} = \{X_1,...\,, X_n\}$ be a set of random variables represented as nodes on a graph $\mathcal{G}$. All the nodes with an outgoing edge directed towards a variable $X$ in the graph  constitute the parent set $\textrm{Pa}(X)$ of $X$. A BN \citep{bnop, probabilistic} is defined as a pair $\langle \mathcal{G}, P \rangle$ of a DAG $\mathcal{G}$ and a joint probability distribution $P$, where $P$ factorizes according to $\mathcal{G}$ into a product of conditional probability distributions of each node given its parents:
\begin{equation}
    P(\textbf{X}) \,=\, \prod_{i=1}^n \, p(X_i|\textrm{Pa}(X_i))\,.
\end{equation}
If the graph $\mathcal{G}$ reflects all and only the conditional independence relationships holding in the distribution $P$, then we say that $P$ and $\mathcal{G}$ are faithful to each other.}

Structure learning in BNs refers to the task of estimating their underlying graph $\mathcal{G}$ from a collection of observed realizations of the random vector $\textbf{X} = \{X_1,...\,, X_n\}$. The number of all possible DAGs grows super-exponentially with the number of nodes $n$ \citep{robinson}, and the problem of recovering the network structure from observational data is NP-hard \citep{chickering2004large}, possibly also due to the acyclicity constraint. Given its practical relevance and the computational challenge, structure learning of Bayesian networks continues to attract interest, with new algorithms constantly developed. For a comparative study and overview of well-established algorithms and some more recent implementations, we refer the reader to \cite{larges} and \cite{benchpress}.

The algorithms for structure learning of BNs fall under two main categories: constraint- or score-based. Constraint-based methods employ conditional independence tests to determine the presence or absence of an edge between each pair of nodes in the network structure. Score-based methods assign a global score to each network to quantify their overall ability to describe the data and search the space of all structures to find high-scoring networks. In this work, we focus on constraint-based methods. 

Since a DAG entails a set of conditional independence relationships \citep{vstruc}, it is possible to at least partially learn the graphical structure by estimating the conditional independencies holding between elements of $\textbf{X}$ from a collection of its observed realizations. However, different DAGs can encode identical sets of conditional independence relations since the same joint distribution $P$ may factorize according to different DAGs. Such a set of DAGs form a Markov equivalence class, and in practice, the underlying graph of a BN is only identifiable up to its equivalence class. 

A completed partially directed acyclic graph (CPDAG) commonly describes an equivalence class of DAGs. A CPDAG is a graph with both directed and undirected edges, and it encodes all the conditional independence statements of a Markov equivalence class \citep{cpdag}. Directed edges in a CPDAG exist in every DAG of the Markov equivalence class it represents. For every undirected edge $X_i \,\textrm{---}\, X_j$ in the CPDAG, at least one DAG exists with $X_i \longrightarrow X_j$ and one with $X_i \longleftarrow X_j$ in the equivalence class. Conveniently, CPDAGs uniquely represent a Markov equivalence class, and therefore in the absence of prior information about the graph, the objective of structure learning reduces to recovering the correct CPDAG.

Two DAGs are Markov equivalent if and only if they share the same skeleton and v-structures \citep{equiv}. The skeleton of a DAG $\mathcal{G}$ is an undirected graph over the same set of nodes with an edge between every pair of adjacent nodes in $\mathcal{G}$. V-structures are triples $X,Y,Z$ of nodes oriented in $\mathcal{G}$ as $X \longrightarrow Z \longleftarrow Y$, and where $X,Y$ are not adjacent. Although estimating the (CP)DAG is the ultimate goal of structure learning, the skeleton can often provide insights into the features of interest of the underlying Bayesian network. Furthermore, \review{for constraint-based methods} obtaining the skeleton is usually a prerequisite for learning a (CP)DAG, a more challenging task sensitive to errors committed while estimating the skeleton. Especially in high-dimensions, skeletons can provide a more accessible but still compelling target.

Learning skeletons is also an essential task of hybrid approaches to structure learning which combine constraint- and score-based methods. Hybrid methods \citep{tsama} aim to combine the computational advantage of conditional independence testing with the higher accuracy of score-based methods to improve the overall performance. A hybrid approach usually employs a constraint-based algorithm first to restrict the space of all DAGs via conditional independence testing. A score-based method then runs over the reduced structure space to find high-scoring networks. Analogously, reliable skeletons may also provide a convenient preliminary search space for sampling methods \citep{kuipers2018}, so developing fast and accurate constraint-based algorithms remains a relevant research topic. 

In section \ref{pc}, we briefly review the PC algorithm and conditional independence testing under the assumption of Gaussian data. Section \ref{dual} introduces the \emph{dual PC} algorithm, our novel variation on the scheme, which provides substantial improvements both in terms of accuracy and run-time by \review{additionally} performing complementary (or dual) tests. Finally, section \ref{sims} reports on a comparative evaluation of the PC algorithm and its dual version on simulated data. To ease reproducibility we provide \texttt{R} implementations of the dual PC algorithm for both its standard and stable version\footnote{Code available at \texttt{https://github.com/enricogiudice/dualPC}}\noindent\blfootnote{\new{A shorter version of this work has already appeared as "The Dual PC Algorithm for Structure Learning" in the Proceedings of the 11th International Conference on Probabilistic Graphical Models, PMLR 186:301-312, 2022.}}.

\section{The PC Algorithm}
\label{pc}

The PC algorithm \citep{pc} is one of the most popular constraint-based structure learning methods. It relies on the faithfulness of the probability distribution of the observed variables $\textbf{X}$ to the unknown DAG $\mathcal{G}$ and the absence of latent confounders (causal sufficiency) for the relationships among $\textbf{X}$.
The algorithm proceeds in two phases: first, it estimates the skeleton by performing a series of conditional independence tests between variables. Second, it directs as many edges as possible while preserving compatibility with the pattern of conditional independencies learned in the first phase. The accuracy of the conditional independence tests is the most critical part of the algorithm since both the skeleton learning and directing of edges depend on them.

\new{The first phase of the PC algorithm goes over the conditional independence tests as follows. The procedure starts from a complete undirected graph $\mathcal{G}$, tests all pairs of variables $X,Y$ for marginal independence and deletes the edge connecting them if it fails to reject independence. After going over all pairs of variables, the algorithm moves on to testing first-order conditional independence relations. For every pair of variables $(X, Y)$ adjacent in $\mathcal{G}$ and every other variable $S \in \textbf{X}\setminus \{X,Y\}$, the algorithm tests if $X$ and $Y$ are independent conditionally on $S$. Again, as soon as it finds a variable $S$ conditionally on which it cannot reject the independence of $X$ and $Y$, the algorithm deletes the edge between them and moves on to a new pair. The algorithm then progressively increases the size of the conditioning sets $S$, and it repeats the procedure for every remaining edge until there can be no higher-order conditional independencies which may result in an edge deletion.}

\new{In practice, we can restrict the search for $S$ to the variables adjacent to $X$ or $Y$. Nevertheless, the first phase of the PC algorithm remains the most computationally intensive since the number of possible independence tests increases exponentially with the number of variables. Because of this, we often need to impose a maximum order on the size of the conditioning set for large networks.}

\new{The second phase of the PC algorithm transforms the skeleton from the first phase into a CPDAG by directing as many edges as possible. Every time it deletes an edge in the first phase, the algorithm also saves the conditioning set $S$, for which it could not reject the independence of the two variables. Such a set constitutes a separating set and helps identify potential v-structures in the second phase. For every triple of adjacent variables in the skeleton $X \,\textrm{---}\, Z \,\textrm{---}\, Y$ where $X$ and $Y$ are non-adjacent, the algorithm orients the triple as $X \longrightarrow Z \longleftarrow Y$ if $Z$ does not belong to the separating set of $X$ and $Y$. The result is a partially directed graph, where we can still determine the direction of some edges to avoid conflicts with the existing v-structures \citep{meek}. }

Under the case of jointly Gaussian data, and as long as the faithfulness and causal sufficiency assumptions hold, the PC algorithm enjoys interesting consistency properties even for asymptotically limited data for sparse graphs, and we refer to \citep{consistent} for further details and pseudo-code. As the sample size goes to infinity, it produces the correct CPDAG. The running time of the PC algorithm is, however, worst-case exponential in the number of variables, though it can execute in polynomial time for sparse graphs \citep{robust}. Despite being one of the most common structure learning algorithms due to its consistency guarantees and relatively simple implementation, the PC algorithm is inefficient when applied to high-dimensional datasets such as gene expression data \citep{pcore}. As a result, several methods aim to improve the efficiency of the algorithm \citep{scalable, reducedpc}. However, they either learn local modules of the structures instead of producing an entire CPDAG, thus compromising the structural accuracy, or rely on additional assumptions concerning the DAG structure to ensure consistency.

Another limitation of the PC algorithm, as described above, is that its results are order-dependent, meaning that it may produce different CPDAGs depending on the order of the variables in the dataset. If it removes an edge incorrectly, the neighbouring sets of other nodes will change, leading to potential additional errors. The presence or lack of given edges in the output may thus depend on the order in which the algorithm executes the conditional independence tests. \cite{order} propose a modification to the original algorithm called PC-stable with the property of being order-independent for the skeleton. Instead of deleting an edge as soon as it finds a separating set, the PC-stable algorithm does not delete any edges until it has tested all edges for a given conditioning set size $|S|$. Therefore, for a given size $|S|$, edge deletions do not influence the possible separating sets for other edges, leading to an output whose skeleton does not depend on the ordering of the variables. Since PC-stable does not implement any graph pruning until it moves to the next size of the conditioning set, it needs to carry out more tests than its standard version. This modification results in a longer running time and exacerbates the existing complexity problem.

\subsection{Sample Version}
In this work, we focus on the case of a $N \times n$ data matrix generated from a jointly Gaussian distribution, where $N$ denotes the number of observations and $n$ is the number of variables. Many applications have focused on the Gaussian case due to the availability of conventional testing procedures for conditional independence \citep{review}. Partial correlation extends Pearson's correlation to measure the degree of association between two random variables conditional on a set of other variables. Full-order partial correlation between two variables in a set measures their correlation in a conditional distribution where one holds all other variables in the set fixed \citep{dictionary}. 

Under a multivariate Gaussian distribution, an explicit relationship links the full-order partial correlations to the precision (inverse covariance) matrix entries. Let $P$ be the precision matrix of the random vector $\textbf{X}\hspace{1pt}$, with $P_{ij}$ the element in the $i$-th row and the $j$-th column. Then
\begin{equation}
\label{ppcor}
    \rho_{X_iX_j|\textbf{X}\setminus \{X_i,X_j\}} \,=\, \frac{-P_{ij}}{\sqrt{P_{ii} P_{jj}}}
\end{equation}
which mimics the marginal correlation obtained from the covariance matrix $\Sigma$ 

\begin{equation}
\label{pcor}
    \rho_{X_iX_j} \,=\, \frac{\Sigma_{ij}}{\sqrt{\Sigma_{ii} \Sigma_{jj}}}\,.
\end{equation}

The concept of partial correlation allows for conditioning on any set $S \subseteq \textbf{X}\setminus \{X_i, X_j\}$ of a lower order.
\new{To test whether an estimated partial correlation coefficient $\hat{\rho}_{X_iX_j|S}$ is significantly different from zero, one typically applies Fisher's $z$-transform:
\begin{equation}
    Z_{X_iX_j|S} \,=\, \frac{1}{2} \log{\left(\frac{1+\hat{\rho}_{X_iX_j|S}}{1-\hat{\rho}_{X_iX_j|S}}\right)}.
\end{equation}

A test will then reject the null hypothesis of zero correlation $H_0: \rho_{X_iX_j|S} = 0$ at the significance level $\alpha$ if 
\begin{equation}
    \sqrt{N - |S| - 3} \,\,|Z_{X_iX_j|S}| \,>\, \Phi^{-1} \left(\frac{1-\alpha}{2}\right).
\end{equation}}

\review{For Gaussian data, however, under the null hypothesis the distribution of the following $t$-statistic
\begin{equation}
    \hat{\rho}_{X_iX_j|S}\sqrt{\frac{N - |S| - 2}{1-\hat{\rho}_{X_iX_j|S}^{2}}}
\end{equation}
follows a $t$-distribution with $(N - |S| - 2)$ degrees of freedom, so we can use an exact test instead of the normal approximation to the $z$-transform above.}

In the Gaussian case, a partial correlation coefficient of zero characterizes conditional independence \citep{lauritzen1996}, leading to an efficient way of testing for conditional independencies, easy to implement within the PC algorithm framework.

\section{The Dual PC Algorithm}
\label{dual}
In its established implementation, the PC algorithm tests conditional independencies starting from zero-order (marginal) independence and incrementally moving to higher-order conditioning sets. The strategy is justified because computing partial correlation coefficients for large conditioning sets is generally computationally costlier. Estimating an $\ell$-order partial correlation coefficient analogously to equation (\ref{ppcor}) requires inverting an $(\ell+2) \times (\ell+2)$ covariance matrix. The matrix inversion step has a polynomial time complexity in the number of variables, slowing the overall procedure whenever evaluating high-order conditional independencies is needed. For denser graphs, however, where variables may share a large number of parents, testing high-order partial correlation coefficients will be unavoidable in the skeleton estimation phase. In such cases, the PC algorithm might have to test a large number of subsets before finding one large enough to render a pair of variables conditionally independent. 

\begin{figure}[t!]
\centering
\includegraphics[width=0.5\textwidth]{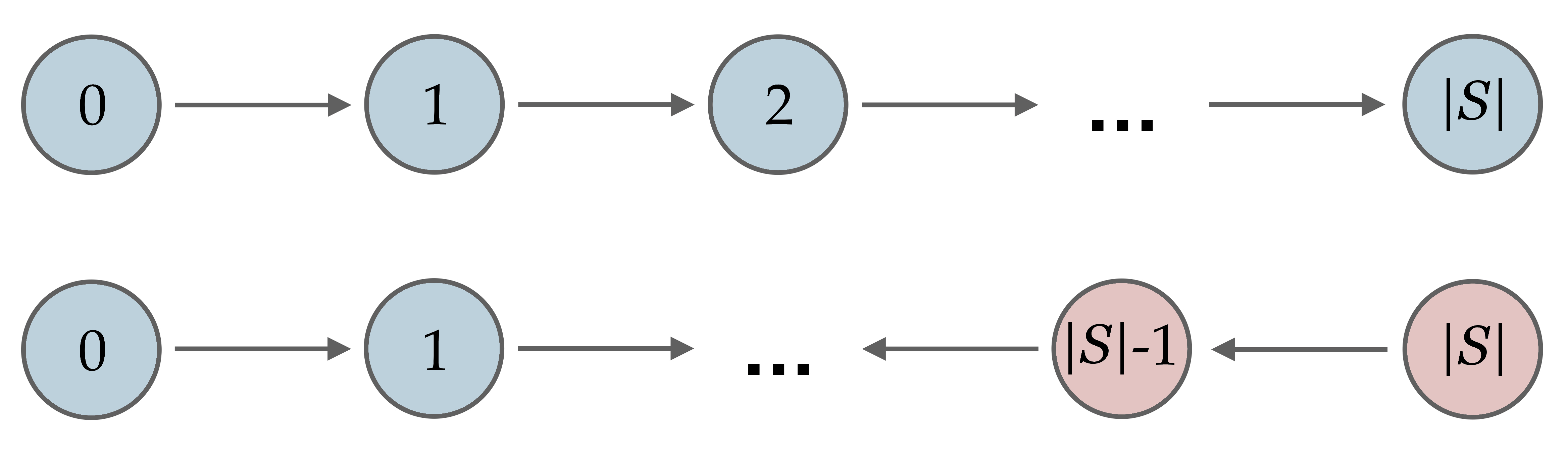}
\caption{The order in which the PC (top) and the dual PC (bottom) algorithms test conditioning set of different sizes for a given pair of variables. $S$ is the current set of variables adjacent to the pair; blue and red sets in the bottom row are complementary to each other. The conditioning set size increases gradually for blue nodes; the dual PC algorithm \review{additionally} performs tests in decreasing order for the red nodes.}
\label{fig:order}
\end{figure}

To overcome the above limitation of the PC algorithm, we propose an alternative ordering of the conditional independence tests, prioritising given high-order partial correlations. The idea is to start testing conditional independence from both zero-order (marginal) and full-order partial correlations among the $n$ variables using the covariance and precision matrices. Our algorithm then proceeds to test more central-order conditioning sets from both directions, starting with first-order and $(|S|-1)$th order partial correlation coefficients, where $S$ is the current set of neighbouring nodes of any pair of variables (see figure \ref{fig:order}). Furthermore, we aim to make the implementation more efficient than the classic PC algorithm by inverting the covariance and precision matrices in blocks and cheaply estimating the partial correlation coefficients.

In the following, we indicate the subset indexed by a set $I$ of a vector $V$ by $V_{[I]}$. If $A$ is a matrix and $I$ and $J$ are two index sets, then we let $A_{[I],[J]}$ denote the $|I| \times |J|$ submatrix of $A$ formed by the entries located in the rows indexed by $I$ and the columns indexed by $J$. For example
\begin{equation*}
    A_{[1,3],[1,3,4]} = \begin{pmatrix} A_{11} & A_{13} & A_{14} \\ A_{31} & A_{33} & A_{34} \end{pmatrix} .
\end{equation*}

Starting with $\Sigma$, the data covariance matrix, our approach first identifies all pairs of variables for which we fail to reject marginal independence, as in a classic PC procedure. After deleting all edges between such pairs of nodes from the initial complete undirected graph, we invert the covariance matrix to obtain the precision matrix. Using the precision matrix, we can further delete any edges between variables for which we fail to reject full-order conditional independence. 

Next, we test pairwise independencies when conditioning on sets of size $1$, again as in the classic PC approach. For every ordered pair $X_i,X_j$ of nodes still adjacent in the current skeleton instance, we find the set $S$ of nodes adjacent to $X_i$, excluding $X_j$. Then we build the local covariance matrix $U = \Sigma_{[i,j,\zeta], [i,j,\zeta]}$ and the local precision matrix $T = U^{-1}$, where $\zeta$ is the index set of $S$: $\textbf{X}_\zeta = S$. There are two advantages to computing $T$: the first is that it will allow us to perform the tests for the complement sets of size $|S|-1$ more efficiently. Indeed, the algorithm will typically need to test a large number of such dual sets, and we can reuse $T$ for every test in the same way we reuse the covariance matrix $U$ for testing multiple sets of size $1$. The second is that by applying equation (\ref{ppcor}) to $T$, we can directly obtain the estimated partial correlation coefficient $\hat{\rho}_{X_iX_j|S}$. 

Let $k$ denote the index of variable $X_\mathcal{K}$ within the local covariance matrix $U$: $\zeta_{(k-2)}=$\,\footnotesize$\mathcal{K}\hspace{1pt}$\normalsize. If we reject the independence of $X_i$ and $X_j$ conditionally on the whole set $S$, we test each variable $X_\mathcal{K} \in S$ by inverting $U_{[1,2,k], [1,2,k]}$ and extracting its first $2 \times 2$ submatrix using block inversion: 
\begin{equation}
\label{ublok}
    \big(U_{[1,2,k], [1,2,k]}\big)^{-1}_{[1,2],[1,2]} \,=\, \left(U_{[1,2], [1,2]} - U_{[1,2], [k]}\,U_{kk}^{-1}\,U_{[k], [1,2]}\right)^{-1}.
\end{equation}

To compute the first-order partial correlations for every $X_\mathcal{K} \in S$, we can then use equation (\ref{ppcor}). Noting that applying equation (\ref{ppcor}) to the $2 \times 2$ matrix we obtain in equation (\ref{ublok}) before the inversion yields the same partial correlation coefficient in absolute value, we can avoid its final inversion. Because the sign of the partial correlation coefficient does not matter for our testing purposes, we can apply equation (\ref{ppcor}) directly without the inversion, speeding up the testing procedure. Therefore we only need to compute
\begin{equation}
\label{peqn}
    \matname(U, k) \,=\, U_{[1,2], [1,2]} - U_{[1,2], [k]}\,U_{kk}^{-1}\,U_{[k], [1,2]} \, , \qquad \vert \hat{\rho}_{X_iX_j|X_\mathcal{K}} \vert \,=\,  \left \vert\frac{B_{12}}{\sqrt{B_{11} B_{22}}} \right \vert .
\end{equation}

Every time the algorithm fails to reject the null hypothesis of independence for a variable $X_\mathcal{K}$, it proceeds to test the complementary (or dual) set $S \setminus X_\mathcal{K}$. To do so, one could naively compute the first $2 \times 2$ block of $(U_{[-k], [-k]})^{-1}$, which would require the inversion of a matrix of size $(|S|-1) \times (|S|-1)$ for every variable $X_\mathcal{K}$. It is more efficient instead to write the desired $2 \times 2$ matrix in terms of $T$:
\begin{align}
\label{tblok}
    \big(U_{[-k], [-k]}\big)^{-1}_{[1,2],[1,2]} \,&=\, \left(T_{[-k], [-k]} - T_{[-k], [k]}\,T_{kk}^{-1}\,T_{[k], [-k]}\right)_{[1,2],[1,2]} \nonumber \\
    \,&=\, T_{[1,2], [1,2]} - T_{[1,2], [k]}\,T_{kk}^{-1}\,T_{[k], [1,2]} \,=\, \matname(T, k)\,.
\end{align}

Therefore by simply applying equations (\ref{peqn}) to the matrix $T$ we can compute $\hat{\rho}_{X_iX_j|S\setminus X_\mathcal{K}}$.  As soon as the algorithm finds a variable $X_\mathcal{K}$ or a dual set $S \setminus X_\mathcal{K}$ conditionally on which it fails to reject independence between $X_i$ and $X_j$, it deletes the edge between the two variables, and it moves on to a new pair. After having tested all the remaining edges, the algorithm proceeds to consider conditioning sets $X_{\mathcal{K}}$ of size $2$ and their dual counterparts. The approach for testing conditional independencies does not change for these new sets since equations (\ref{ublok}) and (\ref{tblok}) hold for index sets $k$ of any size. Analogously to the PC algorithm, the size of the conditioning sets $X_{\mathcal{K}}$ progressively increases until there can be no higher-order conditional independencies that would result in deleting an edge.

Because inverting $U_{kk}$ in equation (\ref{ublok}) can be computationally expensive when considering larger sets $X_{\mathcal{K}}$, we proceed instead by computing its Cholesky decomposition $U_{kk} = C'C$, where $C$ is an upper triangular matrix. To solve the linear system $Cx=U_{[k],[1,2]}$ we can then use back substitution; and finally compute the matrix block of interest as
\begin{equation}
\label{peqn.simple}
    \matname(U, k) \,=\,  U_{[1,2], [1,2]} - x'x \, .
\end{equation}

In our implementation, we test the whole set $S$ before testing any of its subsets $X_{\mathcal{K}}$ since we anyway rely on the local precision matrix $T$ for later dual tests. Indeed, if $|\zeta| < 2|k|$ we use the local precision matrix $T$ to avoid inverting the $|k| \times |k|$ matrix $U_{kk}$ in equation (\ref{ublok}). To compute the partial correlation coefficient $\hat{\rho}_{X_iX_j|X_{\mathcal{K}}}$ it is more efficient to set $k$ equal to its complementary set $\{3,...\,,|\zeta|+2\} \setminus k$ in equation (\ref{tblok}).

\new{Algorithm 1 outlines the pseudo-code for the dual PC procedure. For ease of interpretation, we slightly abuse notation by referring to $i,j,k$ and $S$, both as (sets of) variables and (sets of) their corresponding indices. The comments in the pseudo-code refer to which equations we use, e.g. $T\to$ Eq.~(\ref{ppcor}) indicates that we plug $T$ in place of $P$ in equation (\ref{ppcor}). Although the algorithm output does not change whether one uses the covariance or correlation matrices as input, we employ the correlation matrix in our \texttt{R} implementation for efficiency reasons.

\begin{algorithm}
\caption{The dual PC algorithm - Skeleton learning}
\hspace*{\algorithmicindent} \new{\textbf{Input} Covariance matrix $\Sigma$ of the data with $n$ variables\\
\hspace*{\algorithmicindent} \textbf{Output} Skeleton $\mathcal{G}$ over the $n$ variables in the data}
\begin{algorithmic}[1]
\new{\State Form the complete undirected graph $\mathcal{G}$ over the full set of variables.
\State For every pair of variables $i,j$ delete edge $i\textrm{---}j$ if marginal independence cannot be rejected. \Comment{$\Sigma\to$ Eq.~(\ref{pcor})}
\State Compute precision matrix $P = \Sigma^{-1}$.
\State For every pair of variables $i,j$ delete edge $i\textrm{---}j$ if \NoAlgNumber{\quad full-order conditional independence cannot be rejected.} \Comment{$P\to$ Eq.~(\ref{ppcor})}
\State $\ell \gets 0$
\Repeat
    \State $\ell \gets \ell + 1$
    \Repeat
        \State Select an ordered pair $(i,j)$ of nodes that are adjacent in $\mathcal{G}$.
        \State Find neighbourhood $S$ of nodes adjacent to $i$ excluding $j$.
        \State Build local covariance matrix $U = \Sigma_{[i,j,S], [i,j,S]}$.
        \State Compute local precision matrix $T = U^{-1}$.
        \State Test if $i$ and $j$ are conditionally independent given $S$ \Comment{$T\to$ Eq.~(\ref{ppcor})}
        \Repeat
            \State Select $k \subseteq S$ with $|k|=\ell$.
            \If{$i$ and $j$ are not conditionally independent given $k$} \Comment{$U,k\to$ Eq.~(\ref{peqn})} 
                \State Test for conditional independence of $i$ and $j$ given $S \setminus k$ \Comment{$T,k\to$ Eq.~(\ref{peqn})}
            \EndIf
        \Until{A set is found conditionally on which independence of $i$ and $j$ cannot be rejected}
        \State{\quad or all $k \subseteq S$ with $|k|=\ell$ have been tested.}
        \If{A set was found conditionally on which independence of $i$ and $j$ could not be rejected}
            \State Delete edge $i\textrm{---}j$ in $\mathcal{G}$.
        \EndIf
    \Until{All pairs $(i,j)$ of adjacent nodes in $\mathcal{G}$ have been selected.}
\Until{$|S| < \ell$ for every pair of adjacent nodes $(i,j)$ in $\mathcal{G}$.}}
\end{algorithmic}
\end{algorithm}}
As described in the pseudo-code, the core part of the dual PC algorithm outputs an undirected skeleton. To estimate a CPDAG, we need to direct the edges which identify a unique equivalence class as in the PC algorithm, and we can use the same procedure. This edge-orienting phase requires the algorithm to save all the separating sets, one each time it deletes an edge. Analogously to the classic PC procedure, we can use them to orient v-structures and, afterwards, any other edges for which the acyclicity constraints also determines a direction.

As in the original formulation of the PC algorithm, algorithm 1 does not satisfy the order independence property since the tested sets depend on the previous (potentially incorrect) edge deletions. Modifying the dual PC algorithm to achieve order independence for the skeleton as in \cite{order} is straightforward. In its stable (order-independent) version, we only delete edges after all pairs of variables for a given value of the conditioning set size $\ell$ have been tested, and not straightaway for each conditional independence we fail to reject.

\subsection{Consistency}
Under faithfulness of the distribution $P$ to the true DAG $\mathcal{G}$, the classic PC algorithm is pointwise consistent \citep{pc}, i.e. the algorithm constructs the CPDAG corresponding to the equivalence class of $\mathcal{G}$ as the sample size approaches infinity. If $P$ is Gaussian, the estimated covariance matrix converges to its true value, determining, in the given limit, conditional independencies without errors corresponding to the so-called ``population version'' of the algorithm as sampling variability vanishes. Furthermore, \cite{consistent} proved uniform consistency in the Gaussian case for certain sparse high-dimensional graphs under additional assumptions. 

\cite{pc} formulated the original proof of consistency in the context of causal inference, relying on the concept of d-separation \citep{bnop}. For Bayesian networks $\langle \mathcal{G}, P \rangle$ where $P$ is faithful to $\mathcal{G}$, conditional independence of $X_i$ and $X_j$ given $S \subseteq \textbf{X} \setminus \{X_i,X_j\}$ is equivalent to d-separation of the nodes $X_i$ and $X_j$ given the set $S$ \citep{geigerpearl, vermadsep}. Therefore the considerations in \cite[theorem~5.1]{pc} also apply to the dual PC algorithm since the properties of the output graph remain unaffected by the different ordering of the conditional independence tests. The population version of the dual PC algorithm will share the same properties as that of the classic PC algorithm, and under faithfulness of the Gaussian distribution $P$ to the underlying DAG $\mathcal{G}$, it will produce the CPDAG corresponding to the Markov equivalence class of $\mathcal{G}$.

\review{\subsection{High-dimensional sparsity} \label{sec:hdsparse}

One of the advantages of the PC algorithm is its ability to run and be consistent for large networks as long as the neighbourhoods of the network are sparse (lower order) with respect to the sample size $N$, even if the number of nodes $n$ can be larger and grow faster than $N$ \citep{consistent}. The dual PC algorithm, as it starts with full partial correlation testing requires that the effective sample size (ESS) of $(N-n-1)$ be large enough for testing. 

To get around this restriction, we can simply impose that we do not run the dual tests for conditioning subsets $S$ where the $\mathrm{ESS}=(N-\vert S \vert -3)$ is too small. This allows us to still benefit from the dual tests after the initial pruning has reduced the neighbourhoods of some node pairs, without restricting the applicability to only the large sample regime. In the limit where all dual tests are excluded, the dual PC algorithm will reduce to the classical PC version. 

Following the reasoning of \cite{consistent}, however, as long as the largest neighbourhood, and the largest allowed dual conditioning set are both $o(N)$, then asymptotic consistency holds also for the dual PC algorithm. The latter condition permits the size of the dual conditioning sets to be higher order than the DAG neighbourhoods, as long as it remains lower order than the sample size. 
}

\section{Simulation Study}
\label{sims}
To evaluate the performance of the dual PC algorithm, we simulate synthetic data from randomly sampled BNs. For every simulation step, we generate a DAG over $n$ nodes using the function \texttt{randDAG} from the \texttt{R} package \texttt{pcalg} \citep{pcalg}. Specifically, \review{we use the default settings which samples DAGs as triangular adjacency matrices with iid probabilities of edge inclusion in the triangular part. The probability of edge presence can be varied to modify the density of the generated networks in terms of the expected number of parents per node.}

For each DAG, we then sample a large number $N$ of instances for every node as a noisy linear function of its parents:
\begin{equation}
\label{eq.sim}
Y \,= \sum_{i=1}^{|\textrm{Pa}(Y)|} w_i\,\textrm{Pa}_i(Y) + \epsilon \,,\quad~ \epsilon \sim \mathcal{N}(0,1)
\end{equation}
with the weights $w_i$ sampled from a uniform distribution on the interval $(0.4,2)$. After generation, we standardise the data \review{(by shifting each variable to zero mean and rescaling to unit variance)}. To assess how the algorithm behaves for networks of different sizes, we evaluate the performance of the dual PC across DAGs with $50$, $100$, $150$ and $200$ nodes. For every number of nodes $n$, we also consider different scenarios for the number of observations, with $25n$, $50n$ and $100n$ observations of each variable. For every combination of the parameters $n$ and $N$, we generate $100$ DAGs with their corresponding data matrices; algorithm 2 describes the procedure as pseudo-code.

\new{\begin{algorithm}
\begin{algorithmic}[1]
\caption{Simulation scheme}
\new{\For{$n \in \{50, 100, 150, 200\}$}
    \For{$N \in \{25n, 50n, 100n\}$}
        \For{$i \in \{1, ..., 100\}$}
            \State Generate random DAG $\mathcal{G}_i$.
            \State Generate $N \times n$ data matrix $D_i$ from $\mathcal{G}_i$.
            \For{$\alpha \in \{0.002, 0.005, 0.01, 0.02, 0.05, 0.1, 0.15, 0.2, 0.25\}$}
                \State Apply dual PC algorithm to $D_i$ with significance level $\alpha$ and save result.
            \EndFor
        \EndFor
    \EndFor
\EndFor}
\end{algorithmic}
\end{algorithm}}

\new{To generate the DAGs we choose a setting designed to achieve an expected number of parents for each node equal to $d=2$. The traditional PC algorithm is known to work better for sparser graphs, so we repeat all of the simulations outlined in algorithm 2 in a sparser scenario with the expected number of parents set to $d=1.5$. }

\subsection{Performance Metrics}
\label{perf}
To assess the performance of the dual PC algorithm, we compare its estimated structures to the ground truth structures generated across all scenarios of different sample and graph sizes. Since we can only identify DAGs up to their equivalence class, we compare the true CPDAGs with the estimated ones. As a benchmark, we employ the classic PC algorithm from the popular \texttt{R} package \texttt{pcalg}.\new{Since our algorithm differs from the classic PC algorithm only in its skeleton construction phase, we also compare the results on the estimated skeletons. To assess the effect that the different choices in separating sets of the two algorithms have on the estimated graphs, we also compare the pattern graphs, partially directed acyclic graphs where the v-structures are the only directed edges \citep{meek}.

For the comparative study, we consider two performance metrics for the graph structure:
\begin{itemize}
    \item The structural Hamming distance (SHD),
    \item Receiver operating characteristic (ROC)-like curves.
\end{itemize}

The SHD measures the smallest number of edge additions, deletions and reversals needed to convert the estimated graph into the target one. Thus, a smaller SHD indicates a better estimate of the true underlying graph. }

To build the ROC-like curves, we vary the significance level $\alpha$ of the conditional independence tests and compute the number of true positive (TP) and false positive (FP) edges in the estimated graph. The curves we report differ from traditional ROC curves in that we use the true positive edge rate (TPR) and a modified false positive edge rate (FPRp), both defined with respect to the number of positive edges P present in the ground truth graph as
\begin{equation*}
    \textrm{TPR} \coloneqq \frac{\textrm{TP}}{\textrm{P}} \quad\quad~ \textrm{FPRp} \coloneqq \frac{\textrm{FP}}{\textrm{P}}
\end{equation*}
where P indicates the total number of edges present in the true DAG. In the case of DAGs, if we were to take all possible missing edges over a set of nodes as ``negative edges'', we would typically have a very large number of true negatives, leading to relatively small values for the FPR as compared to the TPR. Therefore, to measure false positive edges on a scale comparable to that of TPR (as for the SHD), we consider the modified FPRp defined above. \review{For directed edges, we include errors in their direction as half a FP and half a false negative (FN), so that the $\textrm{SHD} = \textrm{FP} + \textrm{FN}$, hence
\begin{equation*}
    \frac{\textrm{SHD}}{\textrm{P}} =  \textrm{FPRp} + 1 - \textrm{TPR}
\end{equation*}
and we can read off the rescaled SHD as the Manhattan distance from the top left of the ROC-like curves.}

\new{For each value of $\alpha$, we plot the average TPR and FPRp of the graphs estimated from the sampled datasets, resulting in a curve for each algorithm. Unlike SHD, which summarises the concordance between the estimated and target graph in a single number, a ROC curve provides a more informative comparison since each point comprises two components separately characterising the number of correctly and wrongly included edges.

Following the procedure outlined in algorithm 2, we repeat $100$ times the process of generating a data matrix from a BN for every combination of the following parameters:
 \begin{itemize}
     \item Significance level of the conditional independence tests:
    
     $\alpha \in \{0.002, 0.005, 0.01, 0.02, 0.05, 0.1, 0.15, 0.2, 0.25\}$.
     \item Number of nodes:
    
     $n \in \{50, 100, 150, 200\}$.
     \item Sample sizes of the generated datasets:
    
     $N \in \{25n, 50n, 100n\}$.
     \item Expected number of parents for each node in the DAG:
    
     $d \in \{1.5, 2\}$.
 \end{itemize}

We convert every estimated skeleton to a pattern graph and then a CPDAG to perform the comparative analysis separately for each of the three objects. Furthermore, we compare the output of both versions, the standard and the stable (order-independent) versions of the dual PC algorithm, to the corresponding versions of the classic PC algorithm.}

In addition to measuring the average SHD, TPR and FPRp over the $100$ replications, we evaluate the average run-time of each algorithm. Code to reproduce the full simulations in \texttt{R} is available at \texttt{https://github.com/enricogiudice/dualPC}.

\begin{figure}[t!]
\centering
\includegraphics[width=\textwidth]{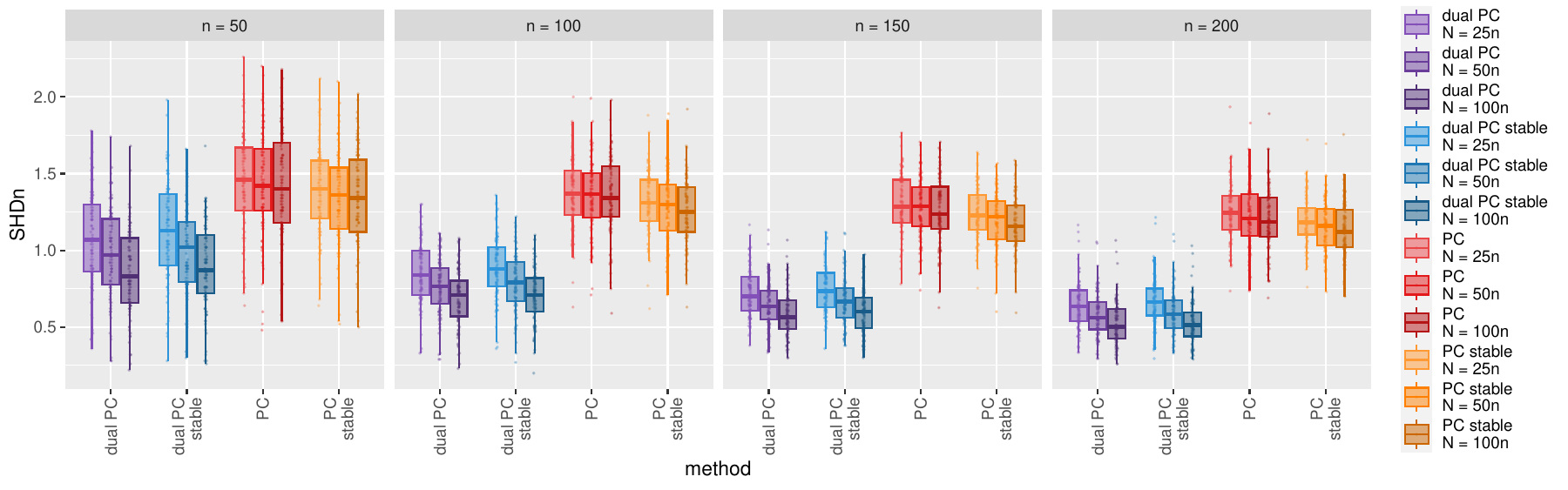}
\caption{\new{Distribution of SHD values scaled by the number of nodes (SHDn = $\mathrm{SHD}/n$) for estimating the correct CPDAG of the PC algorithm, its dual version and their stable counterparts. The significance level $\alpha$ is fixed at $5\%$, and the expected number of parents for a node in the randomly sampled DAGs targets $2$. $N$ indicates the sample sizes of the generated datasets, and $n$ is the number of nodes in each graph.}}
\label{fig:shd_2pa}
\end{figure}

\subsection{Results}
\label{res}
\new{Figure \ref{fig:shd_2pa} shows the distribution of SHD values comparing the estimated and true CPDAGs over alternative scenarios, combining different graph and data sample sizes. For visualisation convenience, we display the SHD values relative to the graph size $n$: $\textrm{SHDn} \coloneqq \textrm{SHD}/n$ so that they are on comparable scales across the different scenarios. The comparison includes four distinct algorithms: the dual PC algorithm and its order-independent counterpart (``dual PC'' and ``dual PC stable''), as well as the standard and stable versions of the PC algorithm. Each method is evaluated in $12$ scenarios, combining different graph and data sample sizes. The significance level $\alpha$ is set at $5\%$. 

The dual PC algorithm performs considerably better than the classic PC, consistently achieving a lower SHD across all simulated settings. Indeed, in every scenario, the dual PC's median SHD is lower than the bottom quartile of the SHD scores of the classic PC algorithm. The relative difference between the two methods becomes increasingly pronounced when the number of nodes in the graphs grows larger.}

\begin{figure}[t!]
\centering
\includegraphics[width=\textwidth]{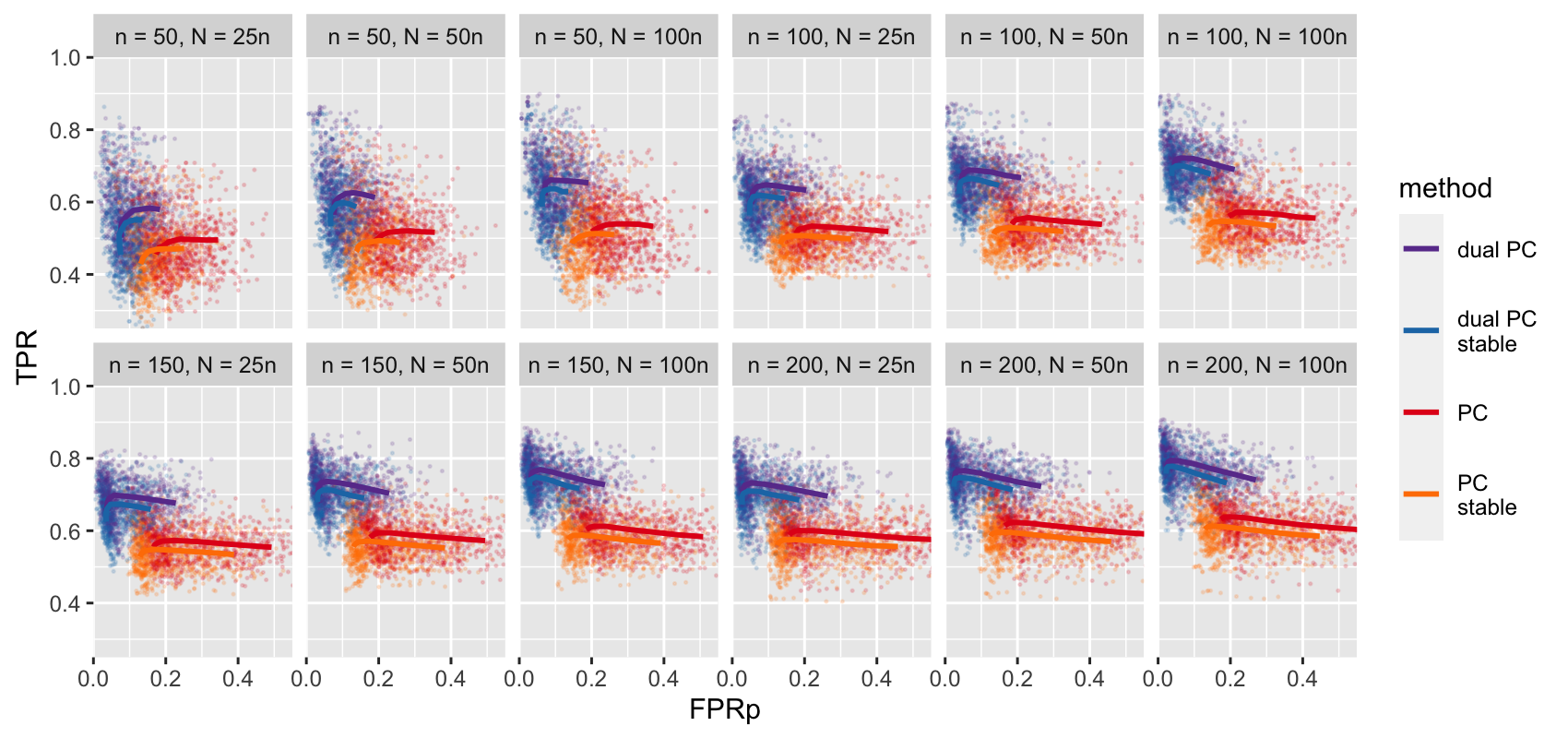}
\caption{ROC-like curves for the dual and classic PC algorithms illustrate each method's ability to recover the correct CPDAG. The software setting in the DAG random sampling targets a value of $2$ for the expected number of parents of each node. $N$ indicates the data sample sizes, and $n$ is the number of nodes in the DAGs.}
\label{fig:cpdag_2pa}
\end{figure}

The scatter plots of figure \ref{fig:cpdag_2pa} display the FPRp and TPR values for different values of the significance level $\alpha$. To calculate the rates, we compare the true and estimated CPDAGs. The solid lines are the (partial) average ROC-like curves constructed by averaging the FPRps and TPRs of every method for each value of $\alpha$. We refer to the lines as \emph{partial} ROC-like curves since they only cover a limited range of false and true positive edge rates determined by the $\alpha$ values considered in algorithm 2. 

\begin{figure}[t!]
\centering
\includegraphics[width=\textwidth]{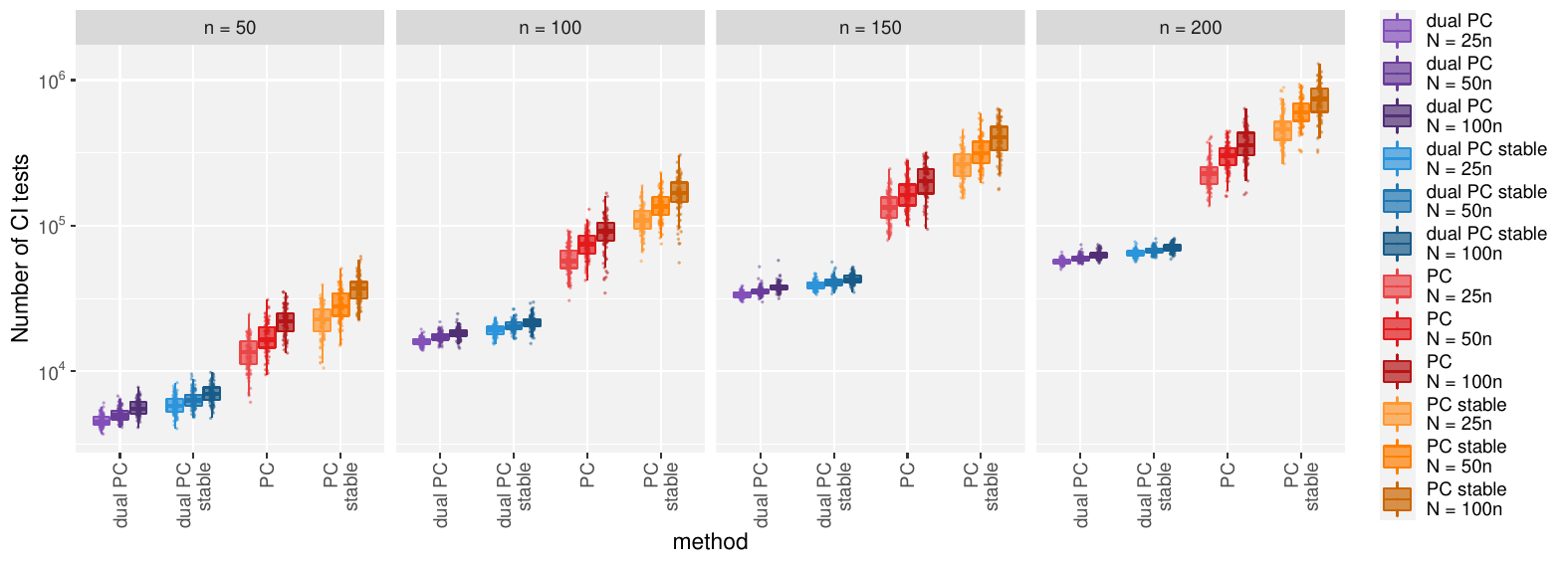}
\caption{\new{Number of conditional independence tests (in log scale) performed by the different algorithms for estimating the skeleton for a significance level $\alpha = 5\%$. The software setting of the function generating random DAGs targets an expected number of parents for each node equal to $2$. $N$ indicates the data sample sizes, and $n$ is the number of nodes in the DAGs.}}
\label{fig:counts_2pa}
\end{figure}

On average, the performance of the dual algorithms is superior in all simulated scenarios, achieving a higher sensitivity for the same level of FPs. For equal values of the significance level $\alpha$, the dual PC offers both: a lower FPRp and higher TPR than the classic PC algorithm. One possible explanation is that the dual PC runs far fewer conditional independence tests, as shown in figure \ref{fig:counts_2pa}). In fact, in the simulated scenarios, the dual version of the PC algorithm performed, on average, at most one-third of the number of tests the classic PC algorithm performed. 

\begin{figure}[t!]
\centering
\includegraphics[width=\textwidth]{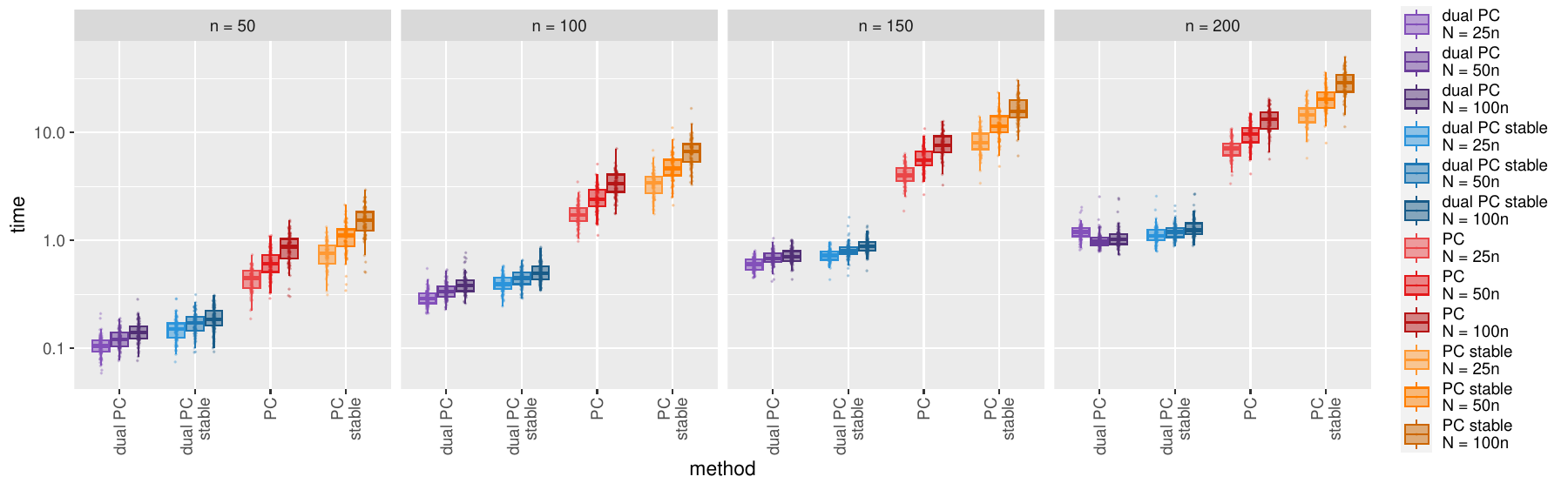}
\caption{Run times for estimating the CPDAG for the considered algorithms for a significance level $\alpha = 5\%$. Time is measured in seconds and displayed on the log scale. The software setting of the function generating random DAGs targets an expected number of parents for each node equal to $2$. $N$ indicates the data sample sizes, and $n$ is the number of nodes in the DAGs.}
\label{fig:time_2pa}
\end{figure}

Figure \ref{fig:time_2pa} shows the distribution of run-times of the different methods in estimating the CPDAG, with the significance level $\alpha$ set to $5\%$. On average, the dual PC algorithm is roughly one order of magnitude faster than the classic implementation of the PC algorithm. The advantage persists across the different combinations of sample and graph sizes. As expected, the stable versions of both algorithms take longer since they require more conditional independence tests. The increased computational cost appears more contained in the dual version than in the classic PC. Again, the reason may be that the dual PC requires fewer tests, while the added computational cost of delaying an edge deletion is not overly burdensome. 

\new{Figures \ref{fig:skeleton_2pa} and \ref{fig:pattern_2pa} in the appendix display the ROC-like curves for skeletons and pattern graphs. Further results in the appendix show that the estimation accuracy and run times for sparser graphs with an expected number of parents of $d=1.5$ agree with the results for $d=2$.

\begin{figure}[t!]
\centering
\includegraphics[width=\textwidth]{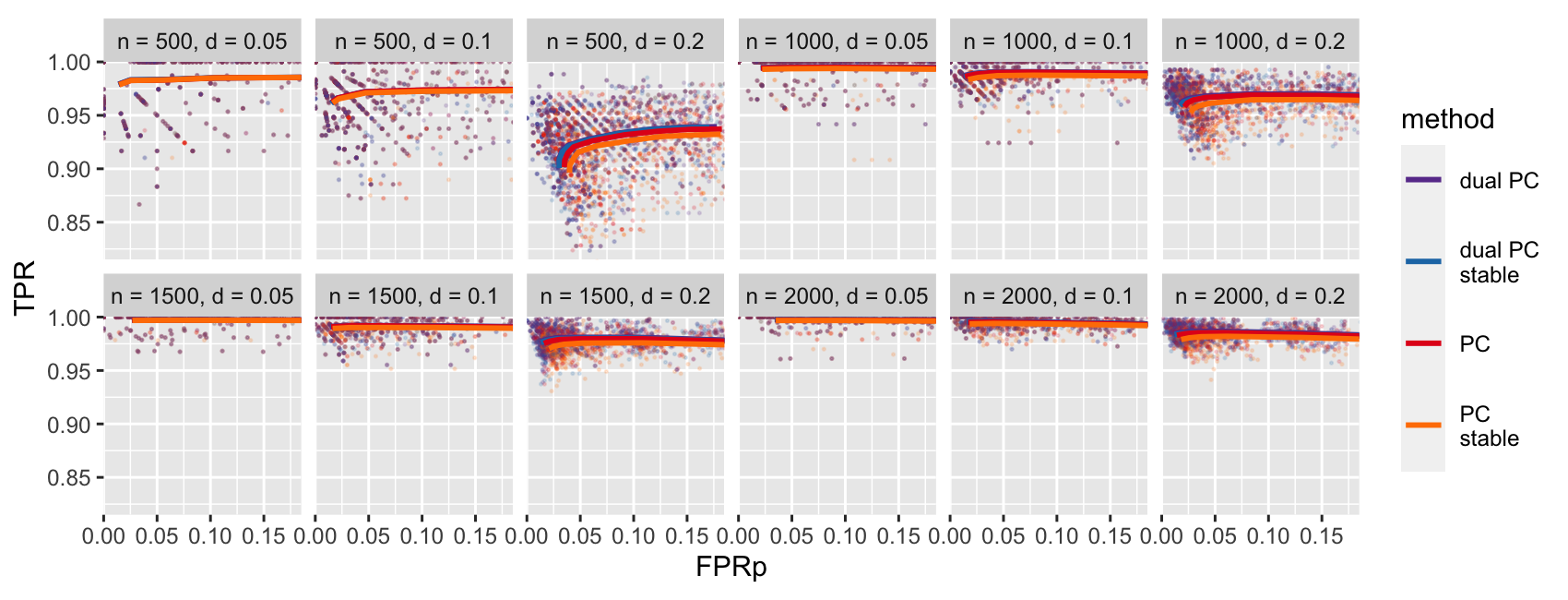}
\caption{\review{ROC-like curves for the ability of the dual and classic PC algorithms to recover the correct CPDAG in the sparse high-dimensional setting with the number of observations just half the dimension.}}
\label{fig:cpdag_sparse}
\end{figure}

\review{\subsection{High-dimensionality}

As discussed in Section~\ref{sec:hdsparse}, we can extend the dual PC to the high-dimensional sparse setting with more variables than observations, by excluding tests without a sufficient ESS. To explore this in simulations, we considered sparse networks with densities of $d \in \{0.05, 0.1, 0.2\}$, but ten times larger than before with sizes $n \in \{500, 1000, 1500, 2000\}$. The sample size was set to $N=\frac{n}{2}$, we correspondingly rescaled the $\alpha$ values by dividing by 1000, and we only ran dual tests if the ESS was at least 20.

The simulation results (Figure~\ref{fig:cpdag_sparse}) show that both the dual and classic PC obtain high TPRs, especially in the sparser settings. They also enjoy very similar performance, with the dual PC having a slight advantage, visible as the density increases. 

Although their accuracy in recovering the networks is very similar, the dual PC version obtains these results more than an order of magnitude more quickly (Figure~\ref{fig:time_sparse}). This speed up is even more pronounced than in the lower-dimensional denser simulations above (Figure~\ref{fig:time_2pa}), and further highlights the advantages of using efficiently-coded dual tests. 

\begin{figure}[t!]
\centering
\includegraphics[width=\textwidth]{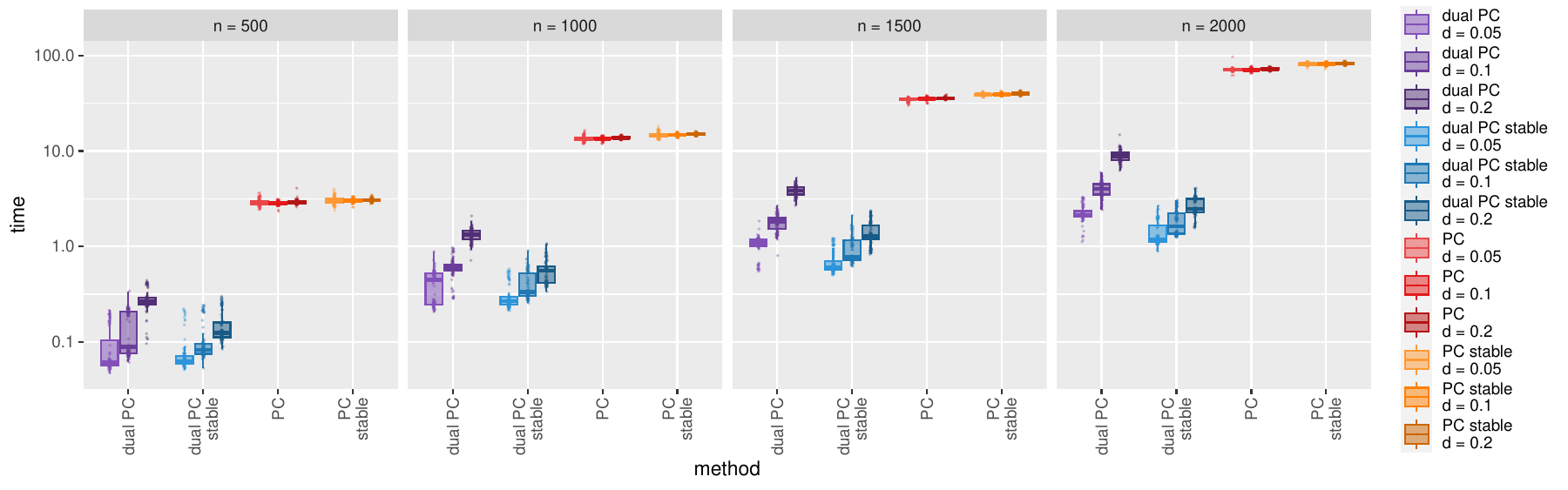}
\caption{Run times (in seconds) for estimating the CPDAG at a significance level $\alpha = 5\times 10^{-5}$ in the sparse high-dimensional setting.}
\label{fig:time_sparse}
\end{figure}

For all the tests so far, we employed the Fisher $z$-transform approximation, which also matches the default in the \texttt{pcalg} package. However, in this sparser high-dimensional setting we may condition on sets for which we have a low ESS.

To explore the effect of using the exact $t$-test instead of the $z$-approximation, we also ran these simulations with the $t$-test. The results are highly concordant between the two tests, but by plotting the difference in SHD (Figure~\ref{fig:SHD_diff} in the Appendix) there is typically the same or a very slight improvement from using the $t$-test since we may benefit from its better calibration and power.
There is also no worsening in the typical run-time (Figure~\ref{fig:time_diff} in the Appendix), as we might expect since transforming the correlation and comparing to the test distribution is not a computational burden.
}

\section{Beyond Gaussianity} \label{beyond_gauss}

The dual PC algorithm relies on the linear Gaussian assumption in two ways. The first is for \review{the exact $t$-test or for the} asymptotic normality of the distribution of the Fisher $z$-transformation of the testing statistic. The central limit theorem implies asymptotic normality of the $z$-transformed partial correlations for any (standardised) bivariate distributions with finite fourth moments \citep{hawkins1989using}. In the absence of Gaussianity, the variance of the asymptotic normal approximation may depend on the value of the correlation coefficient, while there is no dependence for jointly Gaussian data. In the context of the testing implemented in the dual PC algorithm, we are comparing the observed statistic to a null distribution where we assume no correlation. Under such a null distribution, the variance for general bivariate distributions is identical to that for Gaussian data. The Gaussian assumption, therefore, is not critical for the consistency of the tests, though the power may suffer, with larger sample sizes needed to ensure calibration.

Equating zero partial correlation to conditional independence is the second reliance on the linear Gaussian assumption. As expounded in \cite{baba2004partial}, the equivalence between partial correlation and conditional correlation holds for any noise distribution as long as the expected value of the bivariate distribution is a linear combination of the conditioning set. Further, zero partial or conditional correlation is equivalent to conditional independence as long as we can marginally transform the two variables of the conditional distribution into a bivariate normal distribution. Considering the structural equation models for linear DAGs, these conditions are fulfilled for unconnected nodes, even with non-Gaussian noise, if, for example, we only include non-descendants in the conditioning set. For such tests in the dual PC algorithm, the Gaussian assumption is not critical, but as soon as we condition on descendants or if there is a path between the nodes whose independence we are testing, the assumptions break down. Having linearity on any conditioning subset implies that the distribution is multivariate Gaussian \citep{khatri1976characterizations}, hence testing for linearity may be more fundamental than testing for multivariate normality directly \citep{cox1978testing}. In our setting, assuming non-Gaussian noise will, in general terms, affect the linearity needed to test for conditional independence.

To determine how relevant the strict Gaussianity of the data is in practice, we repeat the simulations where the noise $\epsilon$ in equation \ref{eq.sim} follows a Student's t-distribution. \review{This choice allows us to return to Gaussianity by increasing the degrees of freedom $\upnu \to \infty$ and explore stronger violations as $\upnu$ is decreased.} The generated networks have $50$ nodes and $2500$ observations per node. Figure \ref{fig:t_noise_2pa} shows the ROC-like curves under varying $\upnu$ of the noise distribution. The dual PC results are relatively robust to increasing non-Gaussianity, with a slight decrease, particularly in the recall, in line with potentially lower power of the tests and potential misspecification for some conditioning sets as discussed above. Here, we additionally add greedy equivalence search (GES) \citep{ges} as a benchmark. The relative performance of GES compared to the PC algorithm is generally known; GES tends to give more FP edges than PC, but at higher FP rates, it finds more TPs \citep{benchpress}. In terms of finding a skeleton for hybrid methods, the sparser output of the dual PC would generally be preferable since it provides a smaller initial DAG space. The results indicate that deviations from Gaussianity do not substantially affect the performance of the dual PC algorithm relative to the other methods, and it remains competitive. 
\begin{figure}[t!]
\centering
\includegraphics[width=\textwidth]{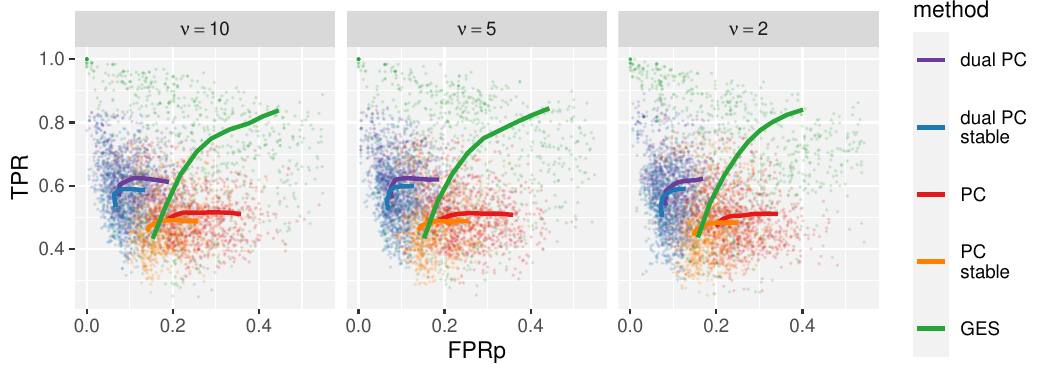}
\caption{\new{ROC-like curves illustrate each method's ability to recover the correct CPDAG for varying levels of non-Gaussianity. For each node, we generated data as a linear function of its parents with Student's t-distributed noise with $\upnu$ degrees of freedom. The software setting in the DAG random sampling targets a value of $2$ for the expected number of parents of each node. See figure \ref{fig:t_noise_15pa} for a similar simulation for sparser networks with $1.5$ parents on average.}}
\label{fig:t_noise_2pa}
\end{figure}

For larger networks (Figure \ref{fig:t_noise_2pa_all}), we observe similar behaviour, but the outperformance of GES over the classic PC algorithm grows with it finding many more TPs at a given level of FPs. On the other hand, the dual PC algorithm does not detect as many TPs as GES, but it can have a much lower FP rate. In all cases, the dual PC algorithm is a clear improvement over the classic version, even in this setting of non-Gaussian noise.}

\new{\subsection{Copula models}

While non-Gaussianity breaks the linearity assumption needed for zero partial correlation to imply conditional independence, a whole class exists of non-linear/non-Gaussian models where the testing procedure of the dual PC algorithm does still apply: nonparanormal or Gaussian copula models \citep{liu2009nonparanormal, liu2012high}.
\review{Specifically, the nonparanormal model assumes that a marginal transformation $f_i(\cdot)$ of each variable $X_i$ exists, such that the transformed variables $f(\textbf{X}) = \{f_1(X_1),...\,, f_n(X_n)\}$ follow a jointly Gaussian distribution. If the transformations defined by the functions $f_i(\cdot)$'s are monotone and differentiable, the nonparanormal distribution is equivalent to a Gaussian copula model. With the freedom of transforming each variable, this kind of modelling provides a richer family of joint distributions for the data than a linear Gaussian model, but where conditional independence statements are still encoded in a latent Gaussian distribution. As for a multivariate Gaussian model, if $\Sigma$ is the covariance matrix of the Gaussian distribution describing $f(\textbf{X})$, the precision matrix $\Omega = \Sigma^{-1}$ also describes the independence relationships of each pair of variables conditional on all other \citep{liu2009nonparanormal}.}

While one traditionally writes copula models in terms of their CDF, we use probability density functions in their nonparanormal form. As shown in \cite{baba2004partial}, if, for the pair of variables whose independence we are testing, a latent space exists where, after a monotonic transformation of each component, they are bivariate normal conditionally on a candidate set, then zero partial correlation implies conditional independence. In a nonparanormal (or Gaussian copula) model, we assume an underlying linear Gaussian model with a marginal transformation applied to each variable, so we are exactly in the setting where, after transforming back to the Gaussian space, the dual PC algorithm is appropriate.

The simplest method for marginally transforming the data back into normal distributions \citep{liu2009nonparanormal, liu2012high} is quantile normalising the data by mapping to the corresponding quantiles of a standard normal. In particular, we map each variable according to
\begin{equation}
\label{eq.qnorm}
Y \to\, \Phi^{-1}\left[\frac{\mathrm{rank}(Y)}{N+1}\right]
\end{equation}
where $\Phi$ is the CDF of a standard normal random variable.

\begin{figure}[t!]
\centering
\includegraphics[width=\textwidth]{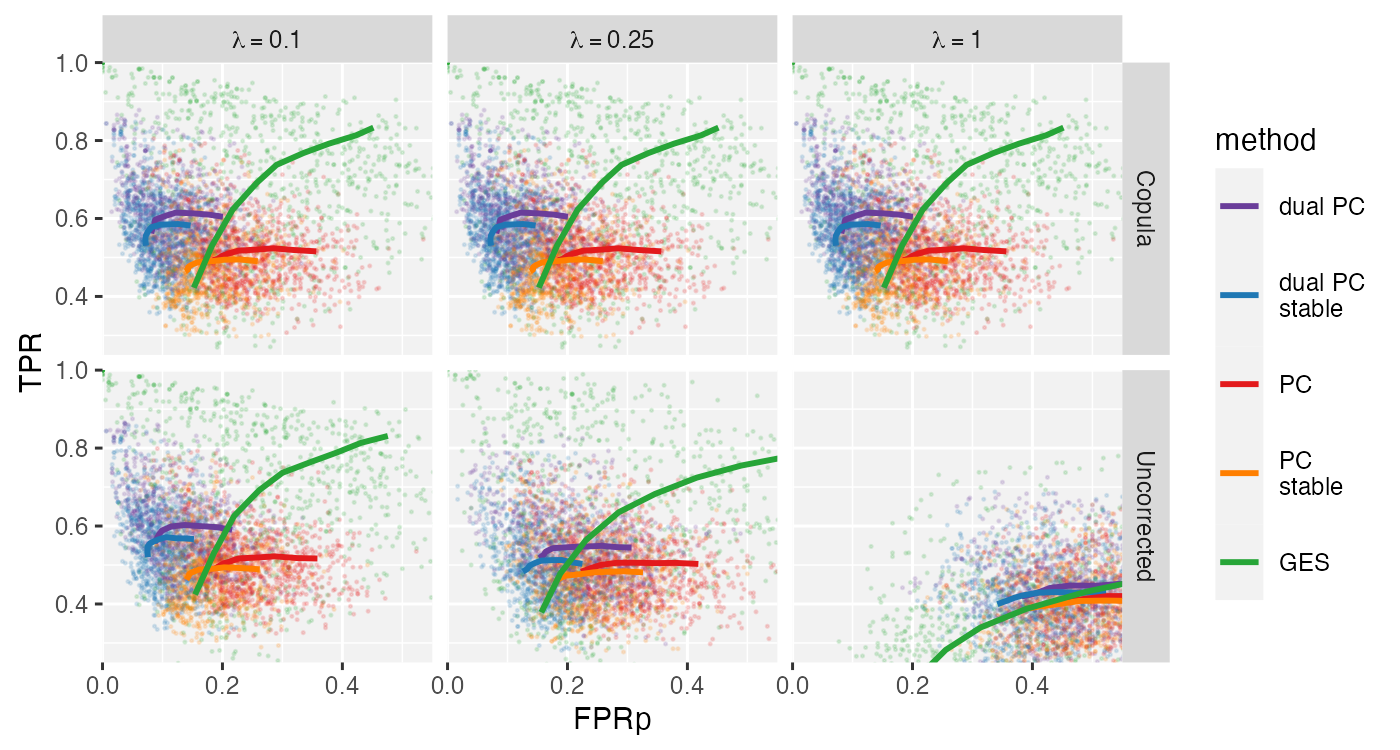}
\caption{\new{ROC-like curves illustrate each method's ability to recover the correct CPDAG for increasing levels of non-linearity of the data transformations, indicated by $\lambda$. On the top row, we perform quantile normalisation, and the copula model is unaffected by the non-linearity while using the uncorrected data leads to a heavy loss of performance. The expected number of parents in the randomly generated DAGs is set to $2$, while the networks have 50 nodes and 2500 observations.}}
\label{fig:copula_2pa_50}
\end{figure}

To test the ability of the dual PC algorithm for copula modes, we first generate standardised linear-Gaussian data as described in section \ref{sims}. Then we pass each variable through the following non-linear transformation:
\begin{equation}
\label{eq.trans}
Y \to \tanh\left[\lambda(Y + \epsilon)\right] \,,\quad~ \epsilon \sim \mathcal{N}(0,1)
\end{equation}
and standardise again. For each variable and each data point, we sample a random shift $\epsilon$ before scaling by $\lambda$ and take the $\tanh$. Higher values of $\lambda$ are more likely to map the data to the tails where the $\tanh$ function is more highly non-linear. In the limit $\lambda\to0$, the transformation leaves the data unchanged.

For the learning, we pass both the uncorrected transformed data to the algorithms and a version of the data after the quantile normalisation of equation (\ref{eq.qnorm}). The results (figure \ref{fig:copula_2pa_50}) show that the performance of the dual PC is completely stable with the quantile normalisation even with increasing levels of non-linearity, as we would expect from the setting with monotonic transformations in both directions and as shared by the other algorithms in the comparison which also use Gaussianity. Notable, however, is that if we run the algorithms directly on the data transformed according to equation (\ref{eq.trans}), the induced non-linearity heavily affects the performance (figure \ref{fig:copula_2pa_50}, second row). This behaviour also holds for larger networks (figures \ref{fig:copula_2pa_100}, \ref{fig:copula_2pa_150} and \ref{fig:copula_2pa_200}).

When the data does follow a latent linear-Gaussian construction, the quantile normalisation in the copula model works well. Its use may, however, degrade performance in other situations. For example, for the data of figure \ref{fig:t_noise_2pa} with Student-t noise, we also run the quantile normalisation and copula construction. The results (top row of figure \ref{fig:t_copula_2pa}) show that using the copula model seems to slightly improve the performance of the classic PC algorithm in the CPDAG reconstruction with increasingly heavy tails of the noise. At the same time GES worsen slightly, and the dual PC algorithm worsen more strongly. With very heavy tails, the dual PC version still outperforms the classic PC version, but the margin is much narrower.}

\section{Conclusions}
This work proposed a novel scheme for running conditional independence tests within the PC algorithm framework. Our algorithm conducts the tests starting \review{both} from zero and full-order conditioning sets and progressively moving to central-order ones from both directions. We harness properties of the precision matrix to compute partial correlations for the dual conditioning sets efficiently. The procedure can perform a larger number of conditional independence tests at a lower computational cost, and it also requires fewer tests altogether so that it enjoys a faster run time than the classic PC algorithm. Importantly, this new approach allows us to efficiently test larger conditioning sets, which the classic version of the PC algorithm would otherwise most likely exclude. 

According to our simulations, the dual PC algorithm achieves a better performance than the classic implementation of the PC algorithm in terms of both SHD, and the ROC like curves displaying TPs and FPs . The advantage persists whether we compare CPDAGs, pattern graphs or skeletons. The run time of our algorithm appears to be much lower on average than the classic implementation of the PC algorithm. Accordingly, the dual PC can efficiently estimate large graphs in high-dimensional settings. 

Bayesian networks provide a natural representation for causally induced conditional independencies and find extensive use for modelling causal relationships between variables \citep{vstruc}. Unsurprisingly, constraint-based structure learning algorithms such as PC are popular tools for learning causal diagrams and effects from observational data \citep{pc, ida}. Improving their speed and accuracy will benefit and extend the possibilities for exploratory causal analyses in high-dimensional settings.

Notably, constrained-based methods constitute an essential component of some hybrid methods for structure learning of Bayesian networks. The PC algorithm may serve to restrict the search space of DAGs for a search and score strategy, where each DAG receives a score, usually a penalized likelihood or a posterior probability. Pruning the DAG space with the help of a constrained-based algorithm before proceeding to search and score may aid exploring the space more efficiently \citep{tsama, kuipers2018}. The dual PC may provide an opportunity to improve hybrid schemes since its increased accuracy and speed compared to the classic PC algorithm would enable a more efficient sampling from the bulk of the posterior probability mass. As such, combining the dual PC algorithm with state-of-the-art MCMC sampling schemes \citep{kuipers2018, gadget} could improve the Bayesian treatment of larger networks.

A limitation of the dual PC algorithm for Gaussian data is that a (partial) correlation coefficient of zero only characterizes (conditional) independence in the jointly Gaussian case,\new{or where there is a latent linear-Gaussian construction as in Gaussian copula models.} On the other hand, the PC algorithm provides a framework to test for conditional independence in more general settings \citep{ordinalPC, copulaPC}. Our simulations show that moderate deviations from Gaussianity do not affect the relative performance of our method, though more general data types may require non-parametric conditional independence testing procedures. Thanks to its versatility, many successfully attempted to extend the PC algorithm for handling non-Gaussian continuous data \citep{genereg, pcdistcor}. In its current version, the dual PC algorithm relies for computational efficiency on the relationship holding under Gaussianity between the precision matrix and the partial correlation. More generally, we might expect improved network learning performance when we supplement low-order conditional independence tests with tests of their complement sets. Our simulations support this by showing that the number of tests necessary for the algorithm to complete is significantly lower for the dual PC than for the classic PC algorithm. Therefore we expect that the reduction in the number of tests might translate into shorter run-times and higher accuracy despite the additional computational burden of testing high-order conditional independencies in the non-Gaussian setting. On these grounds, we speculate that there may be a value in adapting the dual PC algorithm to deal with non-Gaussian\new{(copula)} data, holding the potential to improve the structure learning accuracy and lower the computational cost.

\subsection*{Acknowledgements} The authors are grateful to acknowledge partial funding support for this work from the two Cantons of Basel through project grant PMB-02-18 granted by the ETH Zurich.

\bibliographystyle{apalike2}
\bibliography{references}

\newpage
\appendix
\counterwithin{figure}{section}
\new{\section{Additional Simulation Results}
\label{appe}
Further to the results in section \ref{sims} of the main text, we moreover compare the dual and classic PC algorithms on their ability to recover the correct skeletons and pattern graphs. The corresponding ROC-like curves are displayed in figures \ref{fig:skeleton_2pa} and  \ref{fig:pattern_2pa}; as in the previous plots we construct the curves by averaging the FPRps and TPRs for every value of the significance level $\alpha$.

\begin{figure}[h]
\centering
\includegraphics[width=0.98\textwidth]{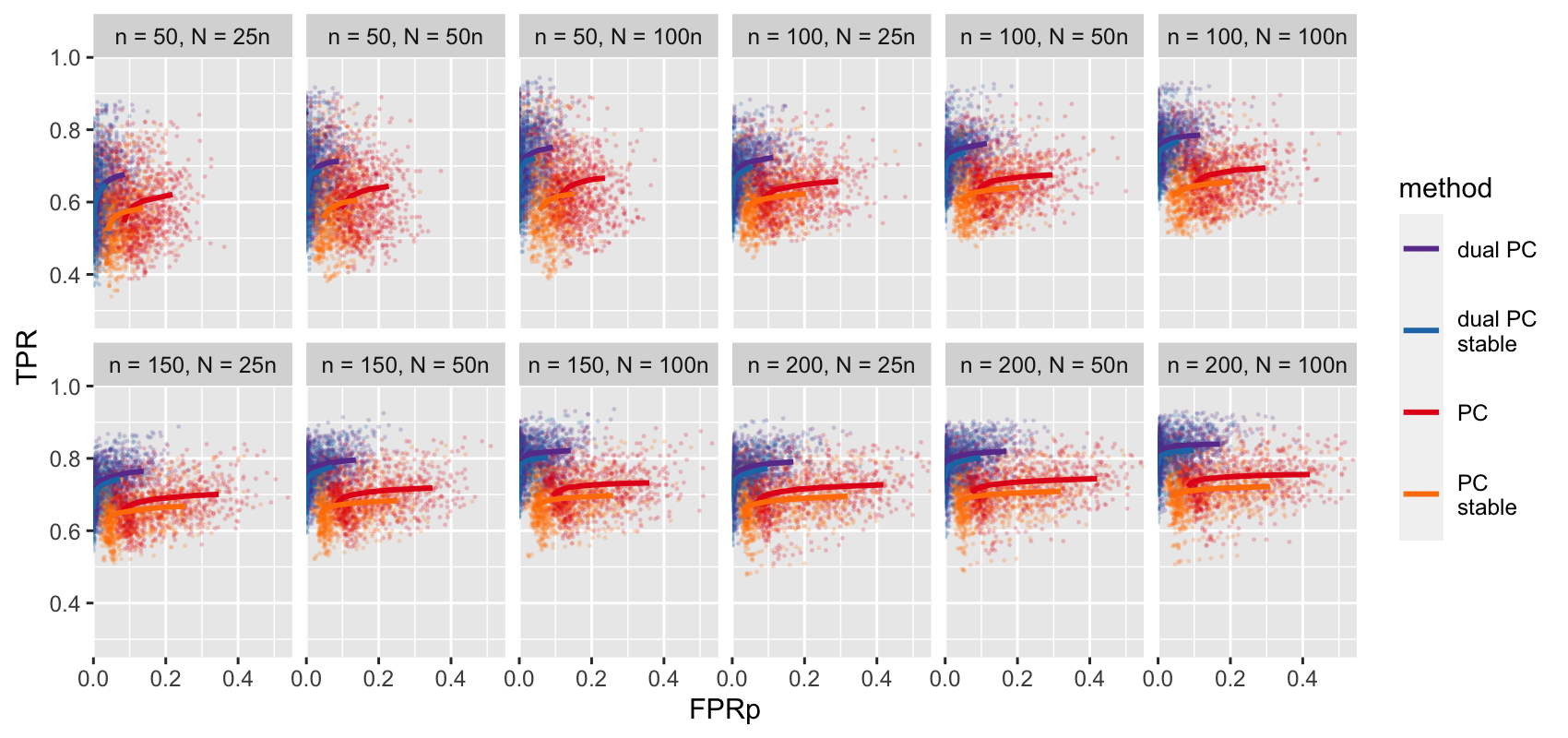}
\caption{\new{ROC-like curves for the dual and classic PC algorithms, illustrating the ability of each method in recovering the correct skeleton. The expected number of parents in the randomly generated DAGs is set to $2$. $N$ indicates the data sample sizes and $n$ the number of nodes in the DAGs.}}
\label{fig:skeleton_2pa}
\end{figure}

\begin{figure}[ht]
\centering
\includegraphics[width=0.98\textwidth]{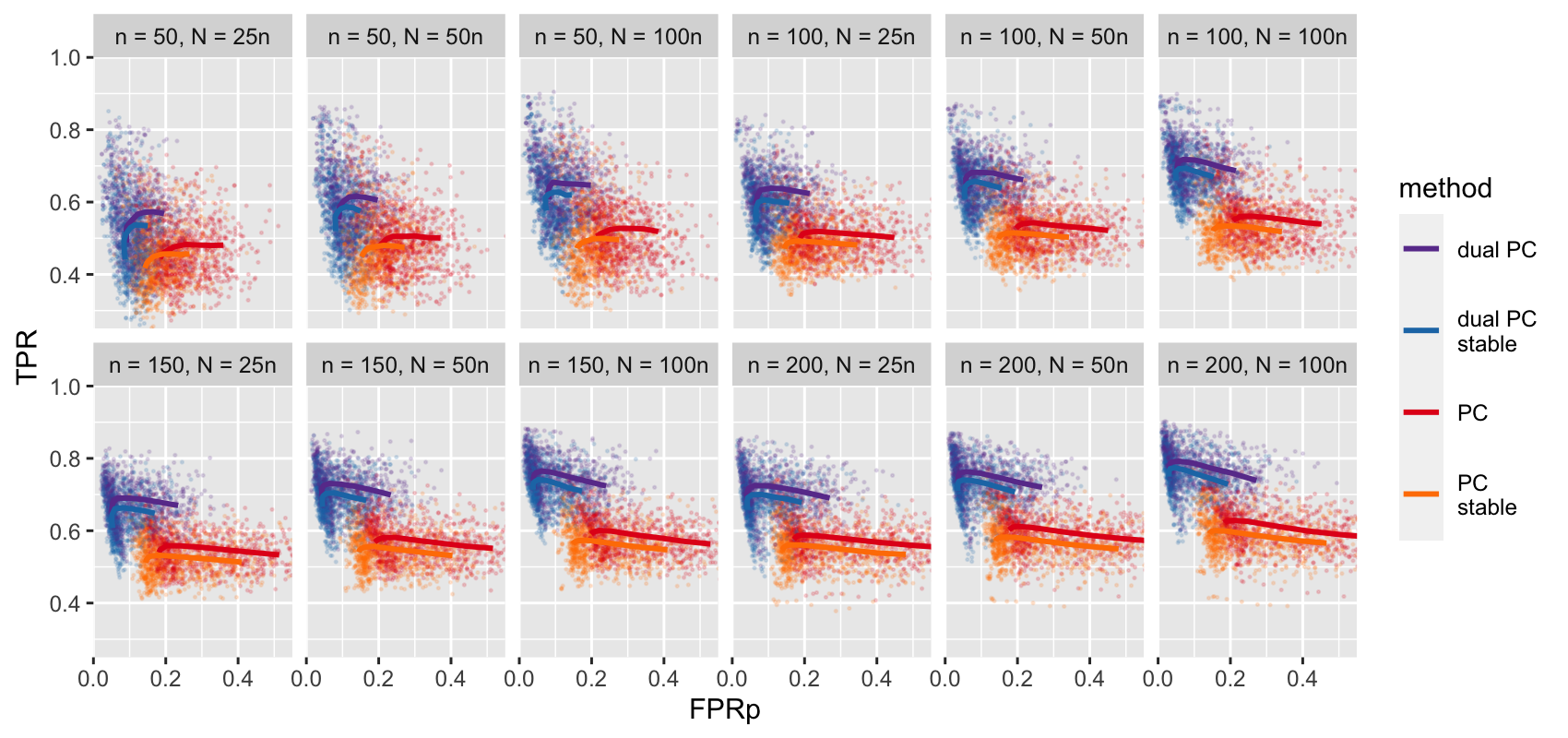}
\caption{\new{ROC-like curves for the dual and classic PC algorithms, illustrating the ability of each method in recovering the correct pattern graph. The expected number of parents in the randomly sampled DAGs is set to $d=2$. $N$ indicates the data sample sizes and $n$ the number of nodes in the DAGs.}}
\label{fig:pattern_2pa}
\end{figure}

To make a comparison on sparser graphs, where the PC algorithm is known to perform well we also consider networks of different density, with values $d=1.5$ for the expected number of parents for each node in the DAG. In sparser scenarios, constraint-based methods typically run faster and are relatively more accurate since a larger number of conditional independence relations hold between variables. Figure \ref{fig:shd_15pa} shows the distribution of SHD values over the $12$ different combinations of sample and graph sizes. Four distinct algorithms are compared in total: the dual PC algorithm and its order-independent counterpart (``dual PC'' and ``dual PC stable''), as well as the standard and stable versions of the PC algorithm. The significance level $\alpha$ is set at $5\%$. The results are in line with those described in section \ref{res}, with the dual PC achieving a lower SHD across the different simulated settings.

\begin{figure}[h]
\centering
\includegraphics[width=\textwidth]{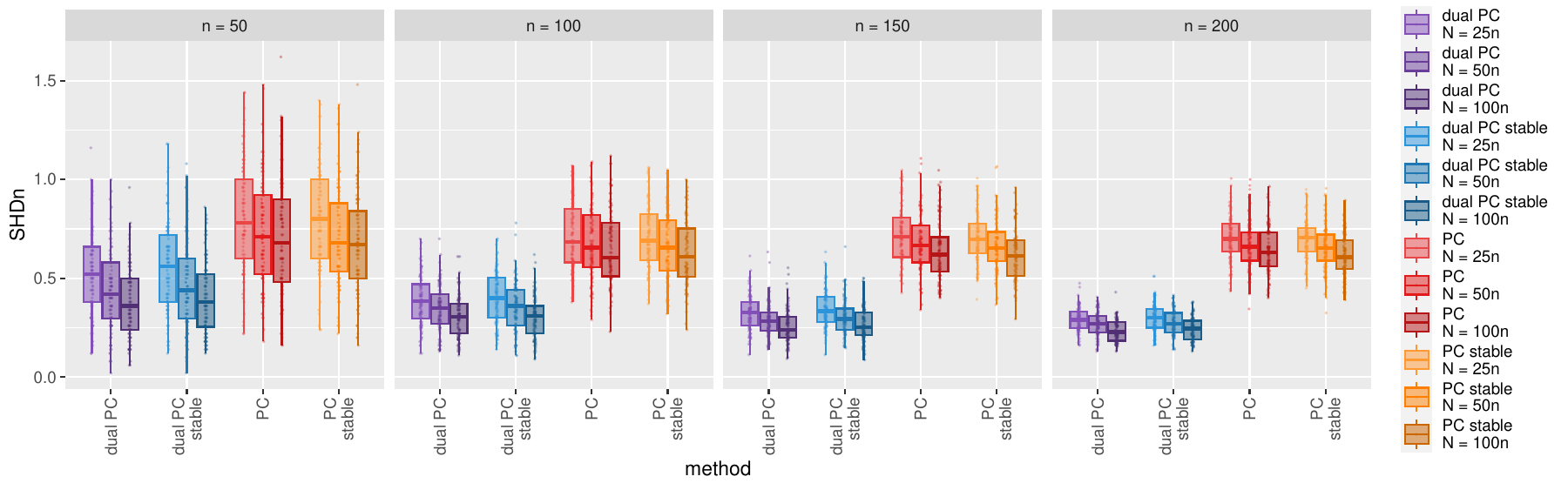}
\caption{\new{Distribution of SHD values scaled by the number of nodes (SHDn = $\mathrm{SHD}/n$) for estimating the correct CPDAG of the PC algorithm, its dual version and their stable counterparts. The significance level $\alpha$ is fixed at $5\%$, and the expected number of parents in the randomly generated DAGs is set to $1.5$. $N$ indicates sample sizes of the generated datasets and $n$ the different number of nodes in each graph.}}
\label{fig:shd_15pa}
\end{figure}

\begin{figure}[h]
\centering
\includegraphics[width=0.98\textwidth]{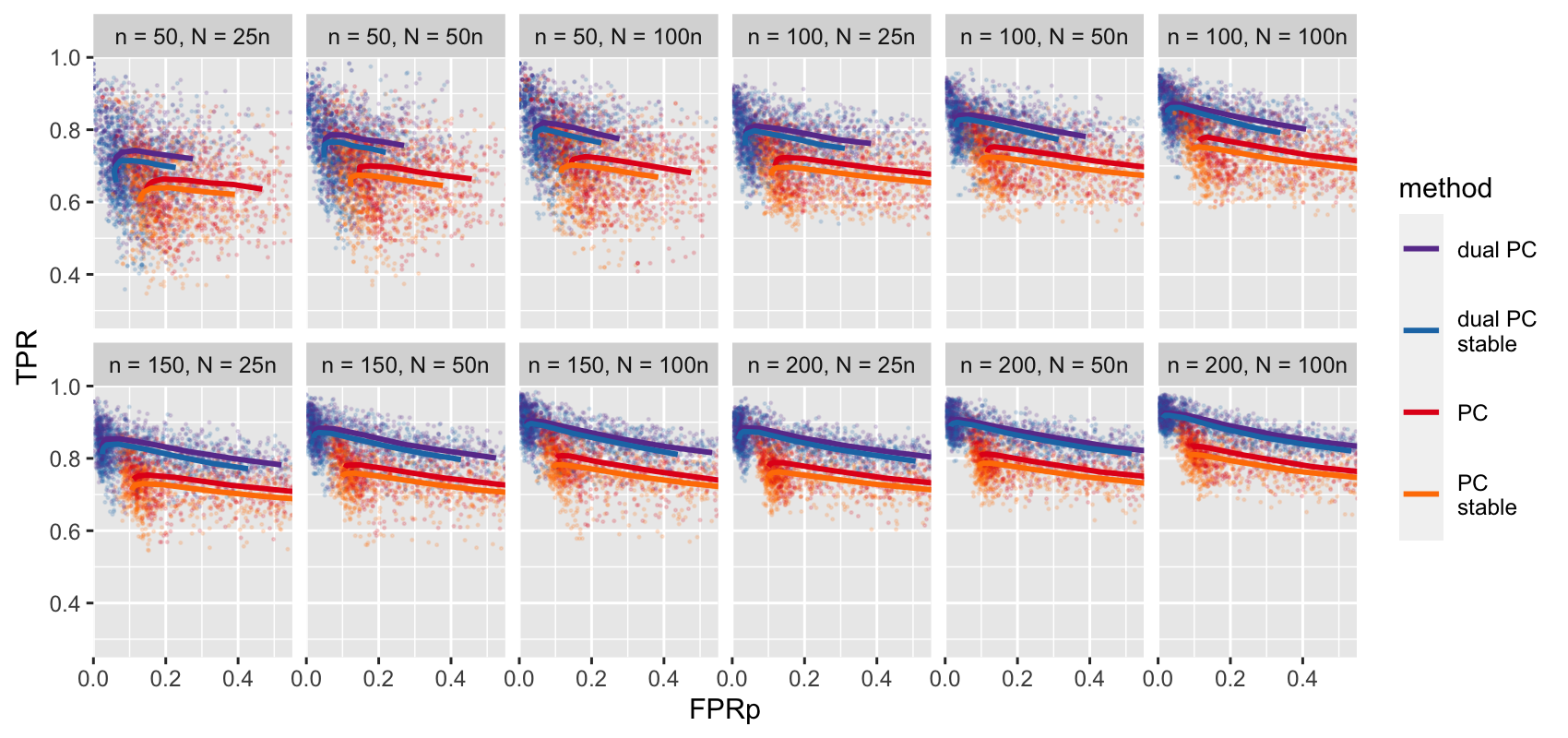}
\caption{\new{ROC-like curves for the dual and classic PC algorithms, illustrating the ability of each method in recovering the correct CPDAG. The expected number of parents in the randomly generated DAGs is set to $1.5$.}}
\label{fig:cpdag_15pa}
\end{figure}

Figures \ref{fig:cpdag_15pa} to \ref{fig:pattern_15pa} display the FPRp and TPR values resulting from different thresholds for the significance level $\alpha$ of the conditional independence tests. The ROC curves are computed by averaging the rates of every method for each value of $\alpha$. The three plots compare the estimated and true CPDAGs, skeletons and pattern graphs, in order. As in the case with the denser graphs, the dual PC algorithm is able to obtain on average both a lower FP and higher TP than the classic PC algorithm for the same value of $\alpha$.

\begin{figure}[h]
\centering
\includegraphics[width=0.98\textwidth]{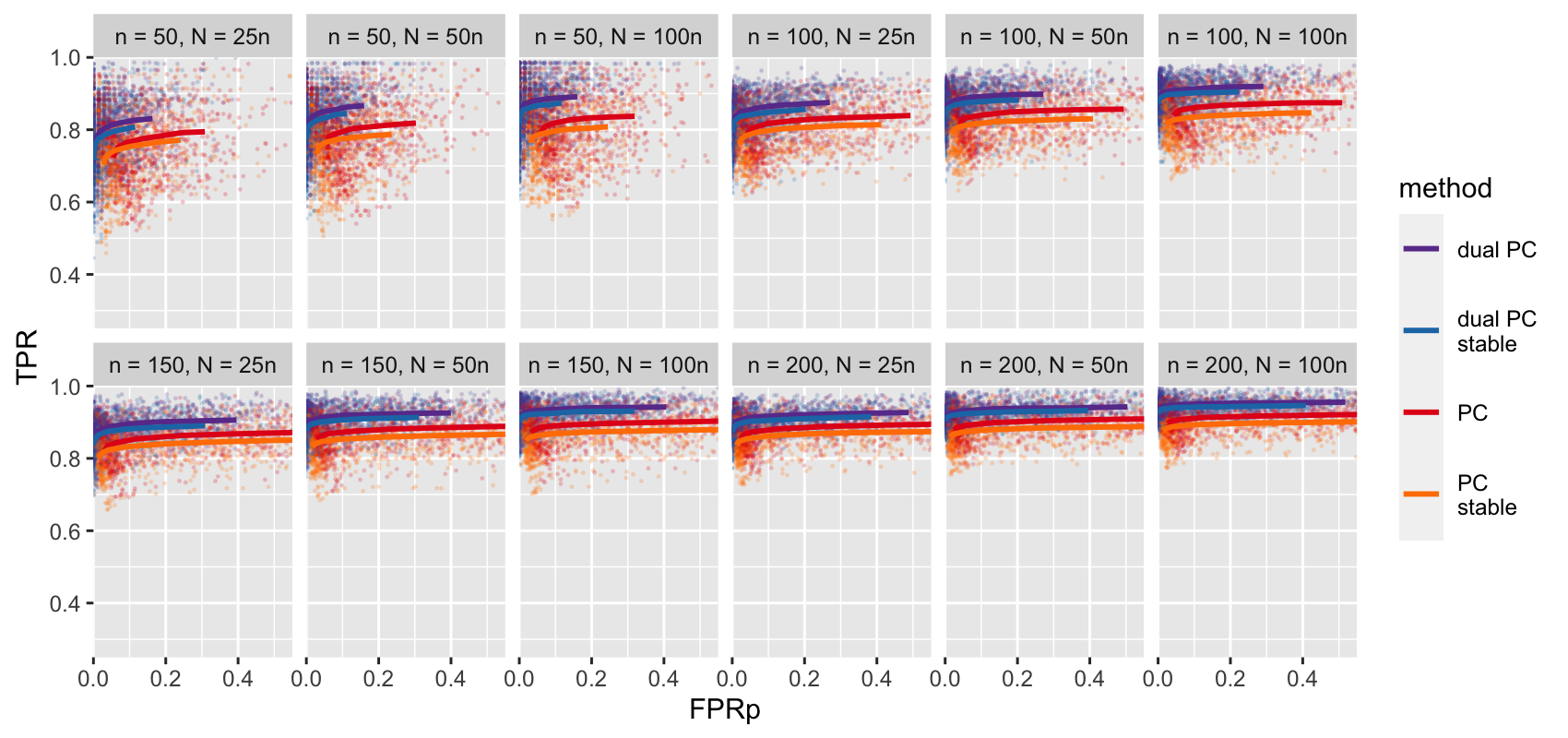}
\caption{\new{ROC-like curves for the dual and classic PC algorithms, illustrating the ability of each method in recovering the correct skeleton. The expected number of parents in the randomly generated DAGs is set to $1.5$.}}
\label{fig:skeleton_15pa}
\end{figure}

\begin{figure}[ht]
\centering
\includegraphics[width=0.98\textwidth]{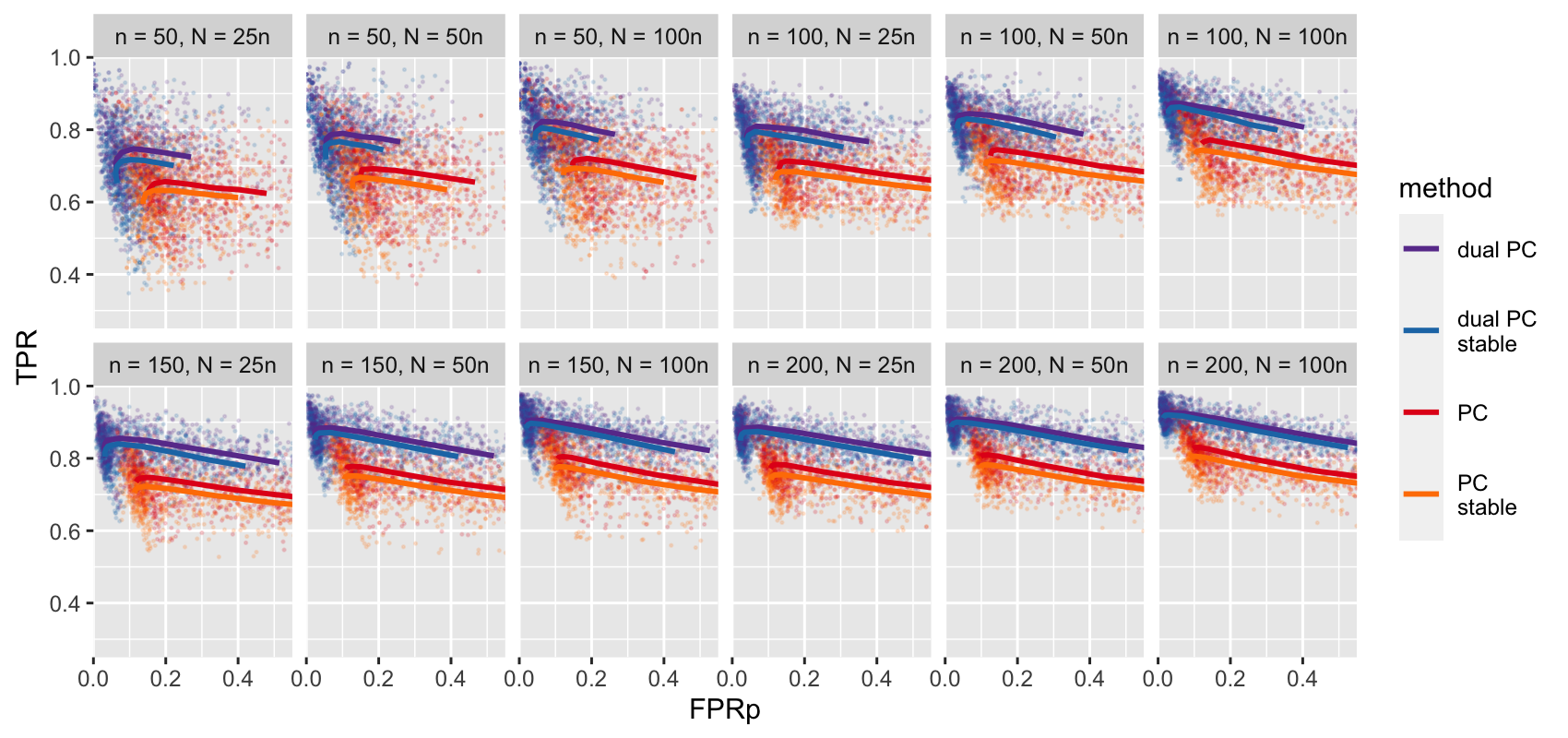}
\caption{\new{ROC-like curves for the dual and classic PC algorithms, illustrating the ability of each method in recovering the correct pattern graph. The expected number of parents in the randomly generated DAGs is set to $1.5$.}}
\label{fig:pattern_15pa}
\end{figure}

Figure \ref{fig:counts_1.5pa} shows the number of conditional independence tests carried out by each of the four considered algorithms. As in the case for $d=2$, the classic PC algorithm exhibits a higher average number of tests than its dual version, especially when comparing the stable versions of both algorithms.

\begin{figure}[ht]
\centering
\includegraphics[width=\textwidth]{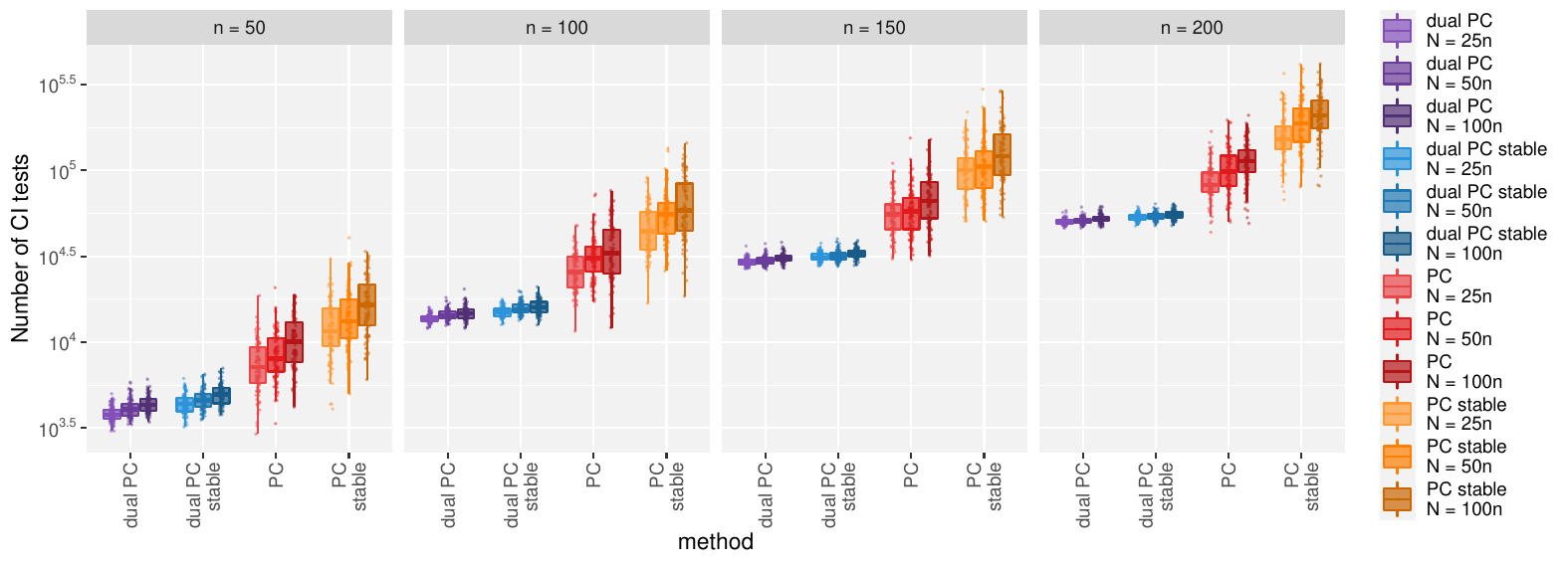}
\caption{\new{Number of conditional independence tests (in log scale) performed by the considered algorithms for estimating the skeleton. The expected number of parents in the randomly generated DAGs is set to $1.5$. $N$ indicates the data sample sizes and $n$ the number of nodes in the DAGs.}}
\label{fig:counts_1.5pa}
\end{figure}

\begin{figure}[ht]
\centering
\includegraphics[width=\textwidth]{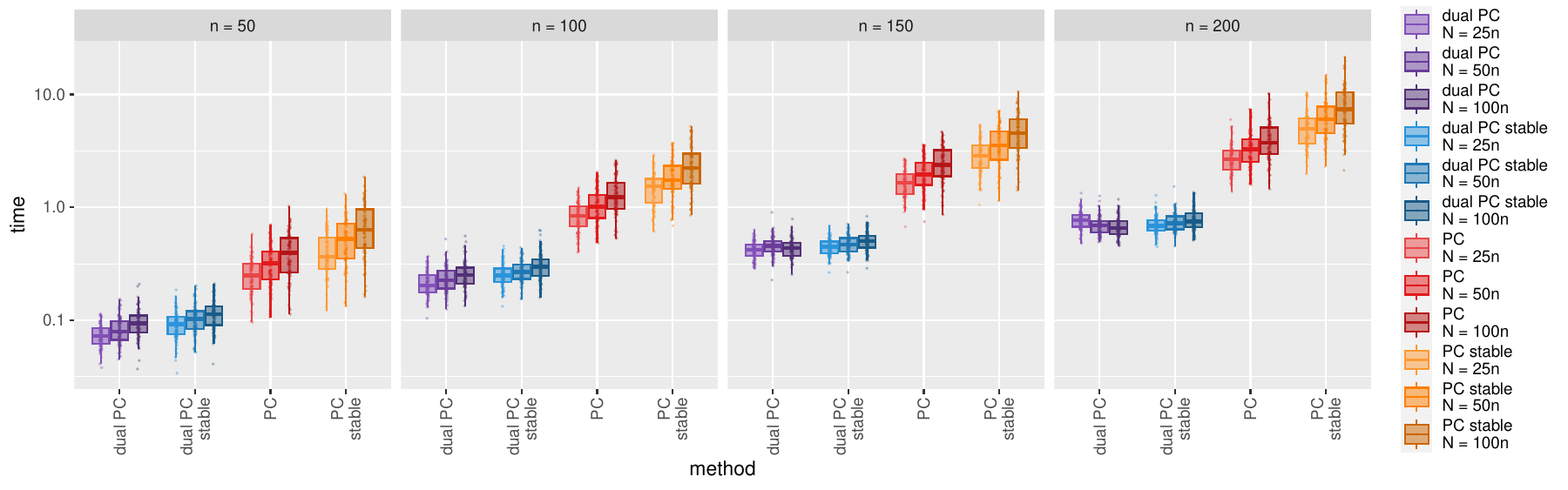}
\caption{\new{Running times for estimating the CPDAG for the considered algorithms. Time is measured in seconds and is displayed in log scale. The expected number of parents in the randomly generated DAGs is set to $1.5$.}}
\label{fig:time_15pa}
\end{figure}

Figure \ref{fig:time_15pa} shows the distribution of run-times of the different methods in estimating the CPDAG, with $\alpha$ set to $5\%$. As expected in this more sparse scenario, both algorithms' run times are shorter compared to the simulations in section \ref{res}. However, the dual PC still maintains a large advantage in speed compared to the classic PC algorithm in each of the simulated settings. Finally, figures \ref{fig:t_noise_15pa} and \ref{fig:t_noise_2pa_all} shows the performance of the dual PC with varying levels of non-Gaussianity, parameterized by the degrees of freedom of a Student's t-distributed noise $\epsilon$ in equation \ref{eq.sim}. Greedy equivalence search is added as an additional benchmark; as discussed in section \ref{beyond_gauss}, the results indicate that the performance of the dual PC relative to the other methods is not greatly affected by deviations from Gaussianity.

\begin{figure}[ht]
\centering
\includegraphics[width=\textwidth]{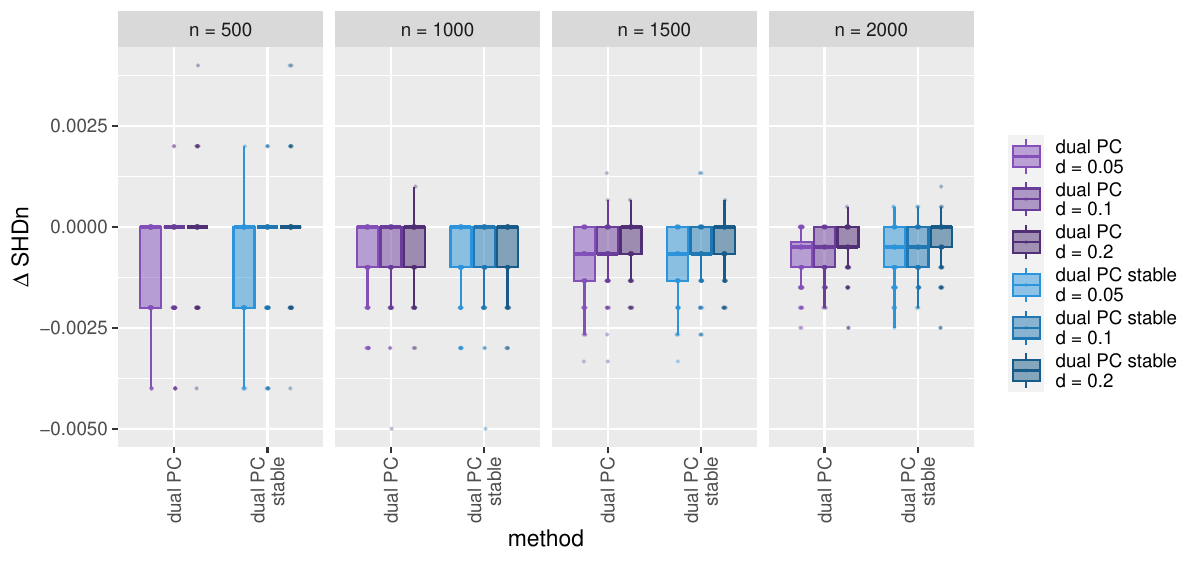}
\caption{\review{Difference in SHD in estimating the CPDAG for the dual PC algorithm when using the exact $t$-test compared to the Fisher $z$-approximation in the sparse high-dimensional setting. The typically lower values for the exact test (negative on the y-axis) indicate slightly better performance with that test.}}
\label{fig:SHD_diff}
\end{figure}

\begin{figure}[ht]
\centering
\includegraphics[width=\textwidth]{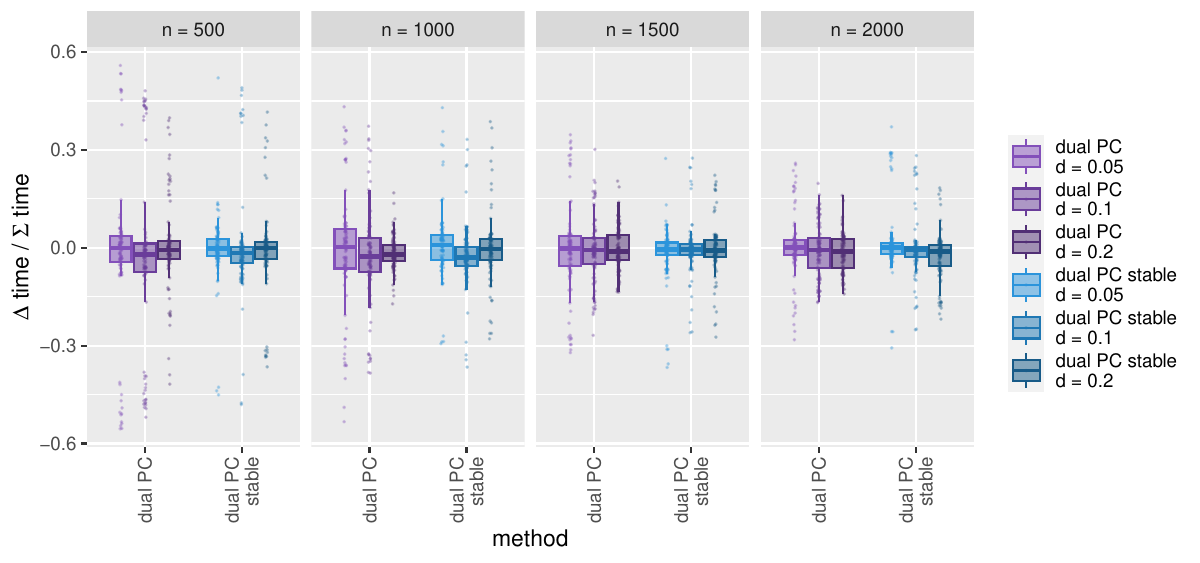}
\caption{\review{Difference in run-times (scaled by the total run-time) for estimating the CPDAG for the dual PC algorithm when using the exact $t$-test compared to the Fisher $z$-approximation in the sparse high-dimensional setting.}}
\label{fig:time_diff}
\end{figure}

\begin{figure}[h]
\centering
\includegraphics[width=\textwidth]{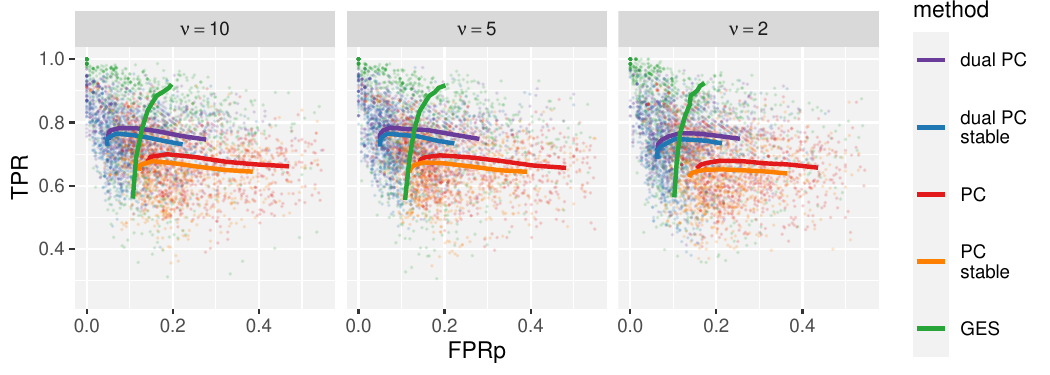}
\caption{\new{ROC-like curves illustrating the ability of different methods in recovering the correct CPDAG for varying levels of non-Gaussianity. The data of each node are generated as a linear function of its parents with Student's t-distributed noise with $\upnu$ degrees of freedom. The expected number of parents in the randomly generated DAGs is set to $1.5$ while the networks have 50 nodes and 2500 observations.}}
\label{fig:t_noise_15pa}
\end{figure}

\begin{figure}[h]
\centering
\includegraphics[width=\textwidth]{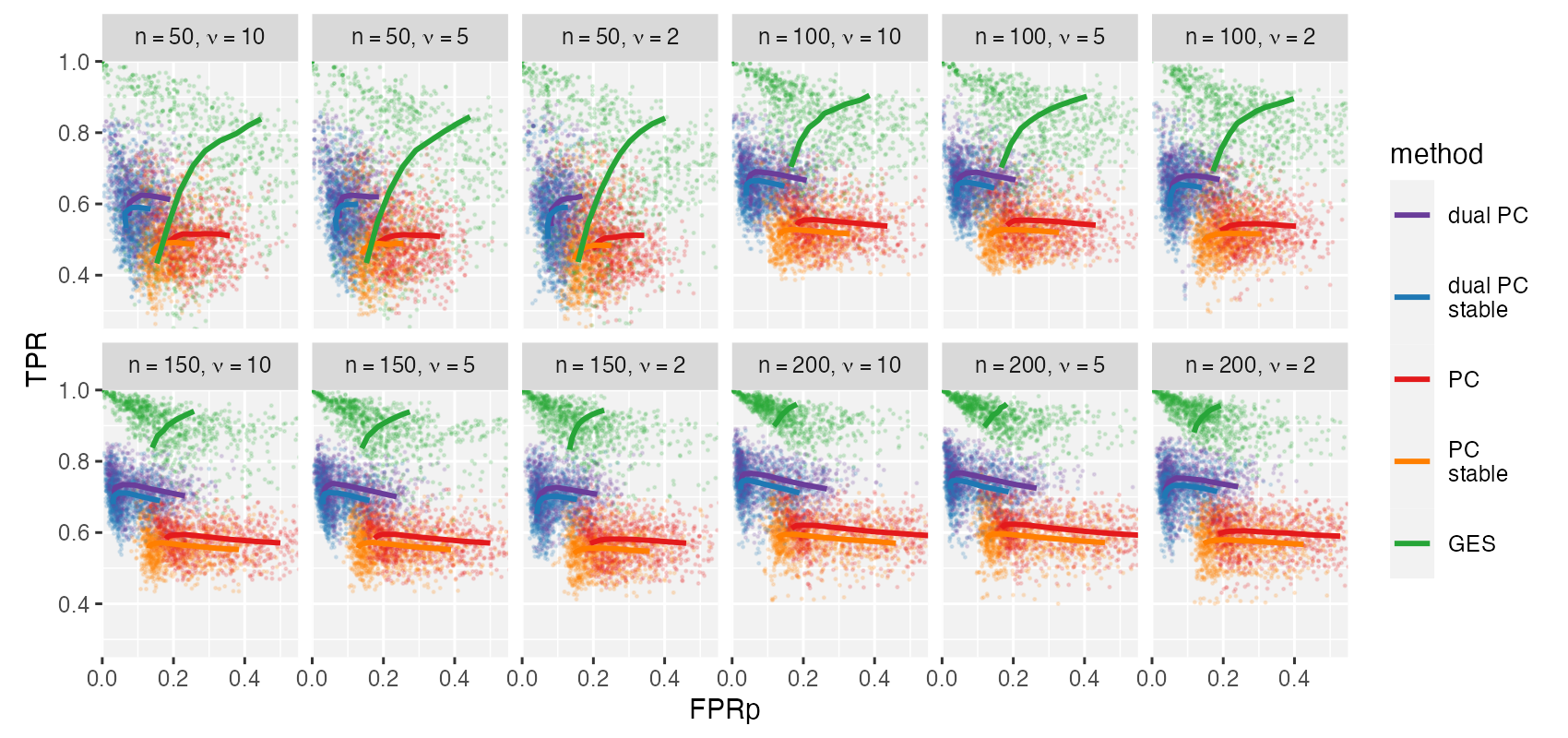}
\caption{\new{ROC-like curves illustrating the ability of different methods in recovering the correct CPDAG for varying levels of non-Gaussianity. The data of each node are generated as a linear function of its parents with Student's t-distributed noise with $\upnu$ degrees of freedom. The expected number of parents in the randomly generated DAGs is set to $2$, while the number of observations $N=50n$ for networks of different sizes $n$.}}
\label{fig:t_noise_2pa_all}
\end{figure}

\begin{figure}[h]
\centering
\includegraphics[width=\textwidth]{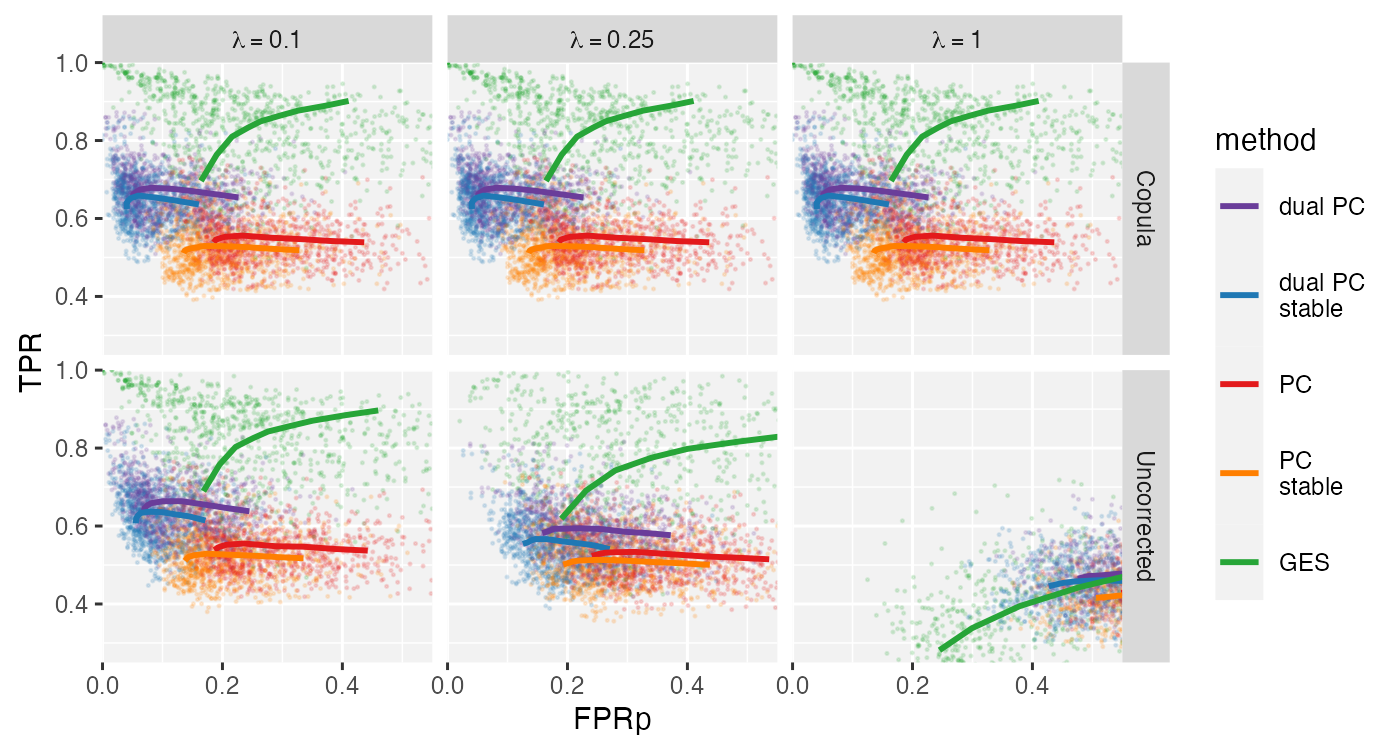}
\caption{\new{ROC-like curves illustrating the ability of different methods in recovering the correct CPDAG for increasing levels of non-linear transformations of the data, indicated by $\lambda$. The expected number of parents in the randomly generated DAGs is set to $2$, while the networks have 100 nodes and 5000 observations.}}
\label{fig:copula_2pa_100}
\end{figure}
\begin{figure}[h]
\centering
\includegraphics[width=\textwidth]{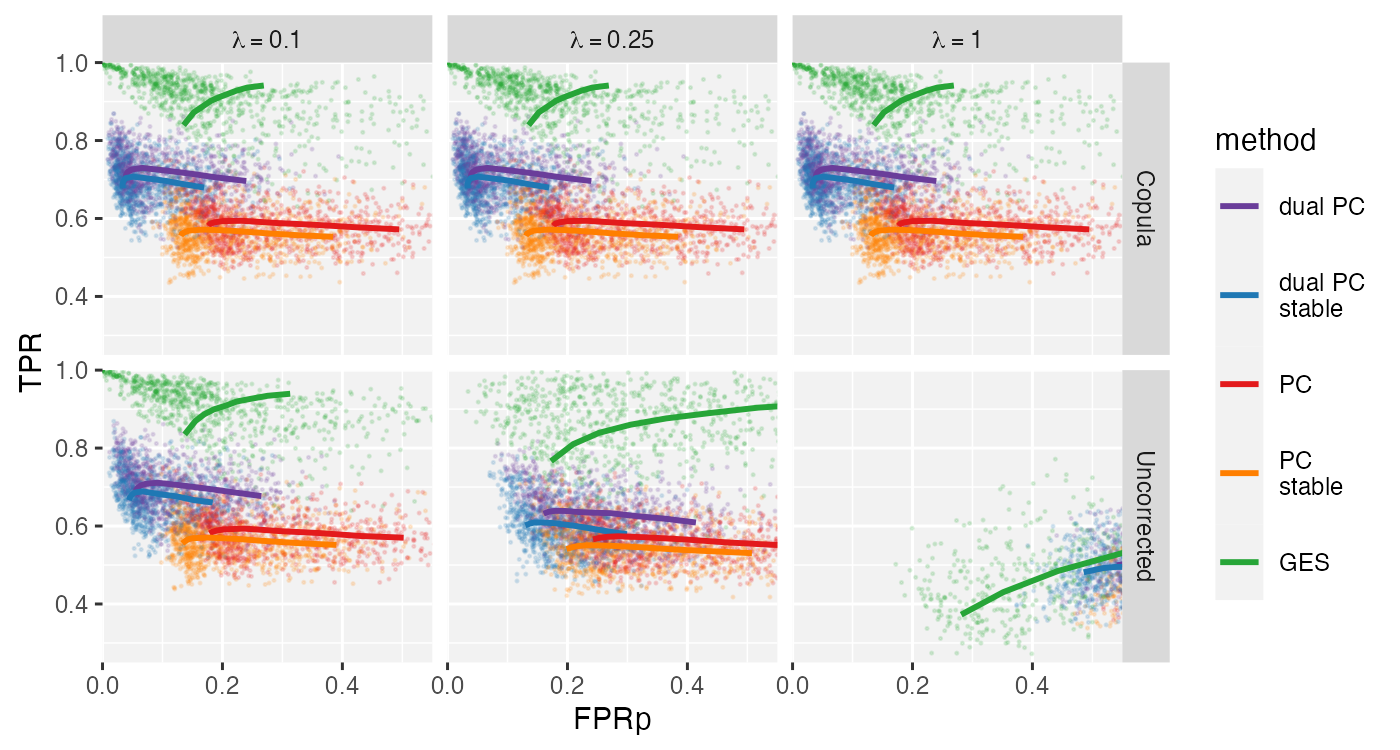}
\caption{\new{ROC-like curves illustrating the ability of different methods in recovering the correct CPDAG for increasing levels of non-linear transformations of the data, indicated by $\lambda$. The expected number of parents in the randomly generated DAGs is set to $2$, while the networks have 150 nodes and 7500 observations.}}
\label{fig:copula_2pa_150}
\end{figure}
\begin{figure}[h]
\centering
\includegraphics[width=\textwidth]{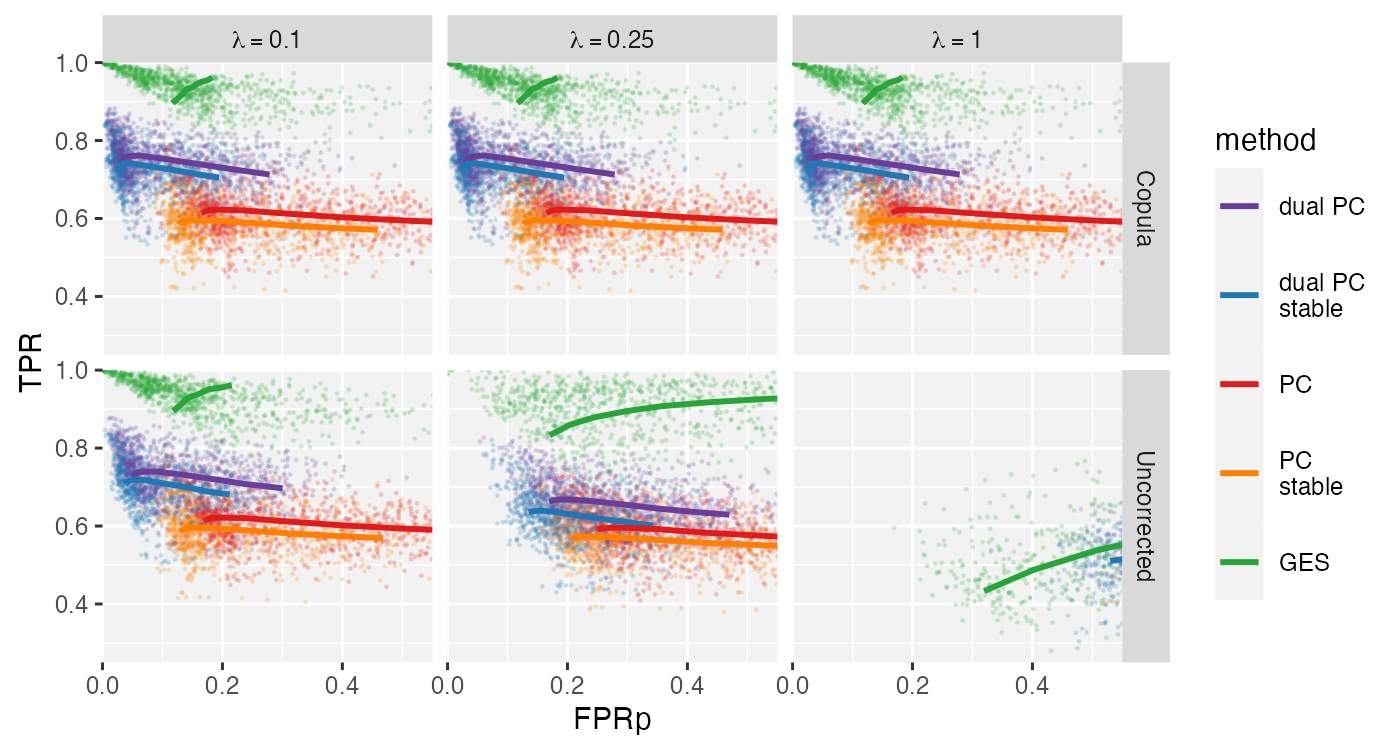}
\caption{\new{ROC-like curves illustrating the ability of different methods in recovering the correct CPDAG for increasing levels of non-linear transformations of the data, indicated by $\lambda$. The expected number of parents in the randomly generated DAGs is set to $2$, while the networks have 200 nodes and 10000 observations.}}
\label{fig:copula_2pa_200}
\end{figure}
\begin{figure}[h]
\centering
\includegraphics[width=\textwidth]{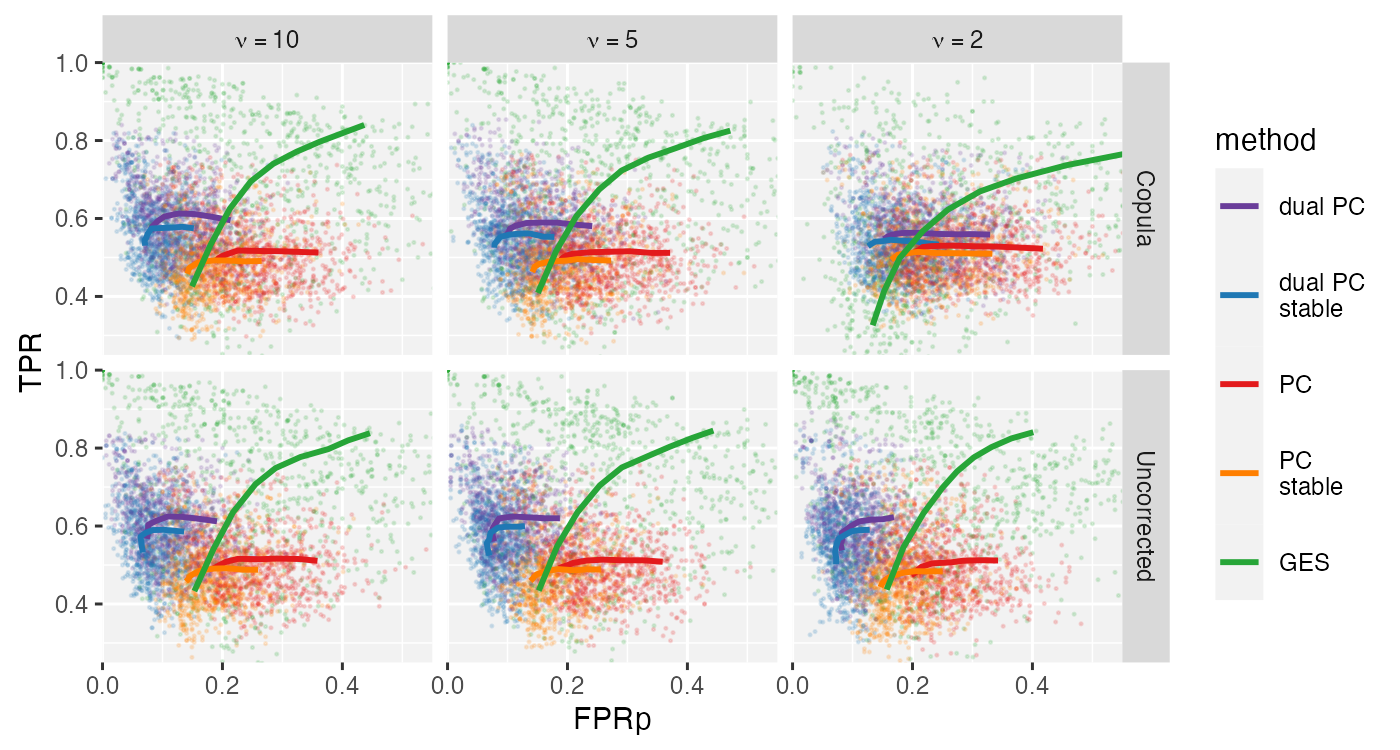}
\caption{\new{ROC-like curves illustrating the ability of different methods in recovering the correct CPDAG for varying levels of non-Gaussianity. The data of each node are generated as a linear function of its parents with Student's t-distributed noise with $\upnu$ degrees of freedom. Along the top row, we employ quantile normalisation to run a copula model, while along the bottom row the data is not transformed giving the results displayed in figure \ref{fig:t_noise_2pa}. The expected number of parents in the randomly generated DAGs is set to $2$, while the networks have 50 nodes and 2500 observations.}}
\label{fig:t_copula_2pa}
\end{figure}
}
\end{document}